%% file: main_camera.tex
\documentclass[10pt,twocolumn,letterpaper]{article}

\usepackage[pagenumbers]{cvpr} % To force page numbers, e.g. for an arXiv version

\usepackage[dvipsnames]{xcolor}
\usepackage{graphics} % for pdf, bitmapped graphics files
\usepackage{epsfig} % for postscript graphics files
\usepackage{times} % assumes new font selection scheme installed
\usepackage{amsmath} % assumes amsmath package installed
\usepackage{amssymb}  % assumes amsmath package installed
\usepackage{enumitem}
\usepackage{mathtools}
\usepackage{algorithm}
\usepackage{algorithmic}
\usepackage{wrapfig}
\usepackage{colortbl}
\usepackage{listings}
\usepackage{afterpage}
\usepackage{relsize}
\usepackage{multirow}
\usepackage{wasysym}
\usepackage{booktabs}
\usepackage{textcomp}
\usepackage{caption}
\usepackage{sidecap}
\usepackage{pifont}
\usepackage[nodisplayskipstretch]{setspace}
\usepackage{xcolor}
\usepackage{url}
\definecolor{nvgreen}{RGB}{118, 185, 0}
\usepackage[pagebackref,breaklinks,colorlinks,allcolors=nvgreen]{hyperref}
\usepackage{fix-cm} % This package lets LaTeX scale the Computer Modern fonts to non-standard sizes.
\usepackage{anyfontsize} % This package is designed to allow more flexible scaling for fonts, which can sometimes help with symbol fonts that don’t have all sizes available.

\definecolor{myblue}{HTML}{3E3EE9}
\definecolor{myorange}{HTML}{FF9000}
\definecolor{mygreen}{HTML}{4B9F2D}
\definecolor{myyellow}{HTML}{D1CA00}
\definecolor{LightGrey}{rgb}{0.92,0.92,0.92}
\definecolor{Myred}{rgb}{1.00,0.12,0.36}
\definecolor{Myblue}{rgb}{0,0.60,0.87}

\newcommand\mypar[1]{\par\vspace{-0.2mm}\noindent\textbf{#1}\;\;}

\newcommand{\ourwork}{Argus\xspace}

\definecolor{cvprblue}{rgb}{0.21,0.49,0.74}

\title{\textsc{Argus}: Vision-Centric Reasoning with Grounded Chain-of-Thought}

\author{
        Yunze Man$^{1,2*}$,
        De-An Huang$^2$,
        Guilin Liu$^2$,
        Shiwei Sheng$^2$,
        Shilong Liu$^{2*}$
    \\
        Liang-Yan Gui$^1$,
        Jan Kautz$^2$,
        Yu-Xiong Wang$^{1\dag}$,
        Zhiding Yu$^{2\dag}$
    \\[0.25cm]
    $^1$University of Illinois Urbana-Champaign~~~~
    $^2$NVIDIA 
}

\begin{document}
\twocolumn[{%
\renewcommand\twocolumn[1][]{#1}%
\maketitle
\begin{center}
    \centering
    \captionsetup{type=figure}
    \vspace{-5mm}
    \includegraphics[trim=0 0 0 0, clip=True, width=0.98\textwidth]{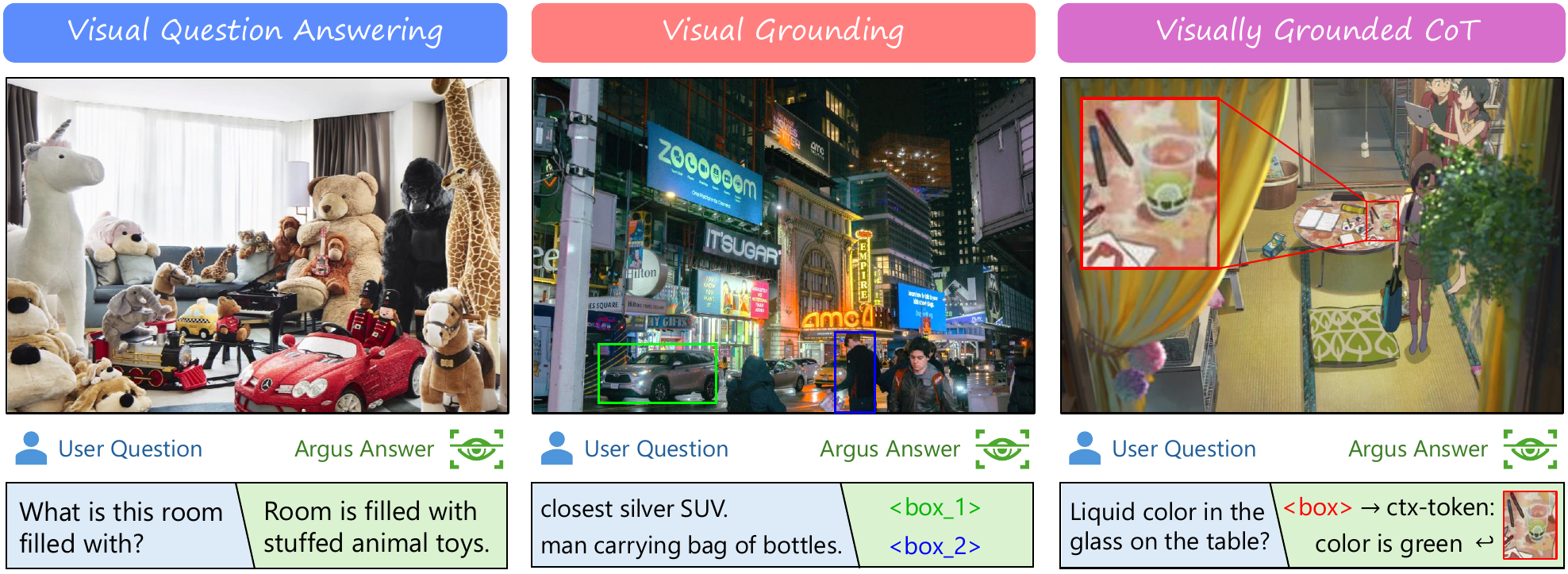}
    \vspace{-3mm}
    \captionof{figure}{Visual question answering, grounding, and chain-of-thought reasoning with {\ourwork}. ``ctx-token'' is short for context token.}
    \label{fig:teaser}
\end{center}}]

\input{sections_ready/0_abstract}   

\input{sections_ready/1_intro}
\input{sections_ready/2_related}

\input{sections_ready/3_method}
\input{sections_ready/4_analysis}

\input{sections_ready/5_conclusion}

{
    \small
    \bibliographystyle{ieeenat_fullname}
    \bibliography{main}
}

\input{sections_ready/X_supplementary}

\end{document}

%% file: sections_ready/0_abstract.tex
\begin{abstract}\vspace{-2mm}
Recent advances in multimodal large language models (MLLMs) have demonstrated remarkable capabilities in vision-language tasks, yet they often struggle with vision-centric scenarios where precise visual focus is needed for accurate reasoning. In this paper, we introduce Argus to address these limitations with a new visual attention grounding mechanism. Our approach employs object-centric grounding as visual chain-of-thought signals, enabling more effective goal-conditioned visual attention during multimodal reasoning tasks. Evaluations on diverse benchmarks demonstrate that Argus excels in both multimodal reasoning tasks and referring object grounding tasks. Extensive analysis further validates various design choices of Argus, and reveals the effectiveness of explicit language-guided visual region-of-interest engagement in MLLMs, highlighting the importance of advancing multimodal intelligence from a visual-centric perspective. Project page: \url{https://yunzeman.github.io/argus/}.
\end{abstract}\vspace{-2mm}

%% file: sections_ready/1_intro.tex
\section{Introduction}
\label{sec:introduction}

Recent breakthroughs in the training of multimodal large language models (MLLMs)~\cite{gpt4v,claude,gemini,llavanext,eagle,cambrian1,minigemini,grok,llavaonevision,vila} have unlocked great advancements in visual-language fusion, allowing these models to extract meaningful content from complex images and perform sophisticated reasoning tasks. However, predominately driven by the success from stronger large language models (LLMs), existing MLLMs still underperform in many vision-centric scenarios~\cite{eyeswideshut,cambrian1,eagle}, where accurate visual perception and understanding determine the success of the subsequent multimodal reasoning tasks (\textit{e.g.,} spatial relationship between objects or properties of specific regions-of-interests (RoIs)). To address these challenges, we \textit{revisit the design space of MLLMs from a vision-centric perspective}, draw insights from cognitive visual intelligence, and propose a new visual attention grounding mechanism for multimodal reasoning tasks, as shown in Figure~\ref{fig:teaser}.

\begin{NoHyper}
\def\thefootnote{*}\footnotetext{Work done during an internship at NVIDIA.}
\def\thefootnote{\dag}\footnotetext{Equal advising.}
\end{NoHyper}

Seminal studies in cognitive science~\cite{naturedirectedandstimulus,james1984psychology} have recognized two distinct types of visual attention mechanism, stimulus-driven visual attention and goal-directed visual attention, which are also referred to as involuntary attention and voluntary attention~\cite{brainmechanisms,directedattention,james1984psychology,morecraft1993architecture}, respectively. Stimulus-driven visual attention is an automatic bottom-up capture of attention driven by salient objects or textures in the visual stimulus. On the other hand, goal-directed attention is a top-down conscious selection of attention, driven by goals and intentions. Surprisingly, an interesting analogy of the two mechanisms of visual attention is presented in the design space of the MLLMs, where (1) the image tokenization with pre-trained visual foundation models~\cite{radford2021clip,kirillov2023segmentanything,oquab2024dinov2} represents the stimulus-driven attention, and (2) the language-conditioned image feature engagement happening within the LLM's transformer layers represents the goal-driven attention. The illustration of the visual attention activation strengths in Figure~\ref{fig:two_attentions} clearly demonstrates different focus areas of two attention modules in MLLMs. 

Although several existing methods~\cite{ferretv2,cambrian1,eagle} have studied and highlighted the importance of unconditioned image tokenization to modern MLLMs' reasoning capacity through knowledge distillation and mixture-of-vision-experts (MoVEs), the effect of explicit language-guided visual engagement is underexplored in the research community. This raises two natural questions: 

1) \textit{What is the best way to introduce a language-directed visual attention mechanism into the design of MLLMs?}

2) \textit{In addition to perception tasks, can a more explicit visual engagement benefit multimodal reasoning tasks?}

To answer these questions, we explore and propose a grounding-driven visual attention re-engagment module in the multimodal causal prediction process. Unlike most existing MLLMs relying on an implicit self-attention mechanism to model language-directed visual token attendance~\cite{llavanext,cambrian1}, we pivot to an \textit{explicit top-down visual search} to locate the image RoI most relevant to the text prompt, and then guide the model to focus on the searched regions for subsequent reasoning and answer generation. Recent work has shown that an object-centric representation benefits the vision-language alignment process and subsequent perception tasks~\cite{contrastivelocalizedlanguageimage,ferret}. Hence, we utilize text-to-box object-centric grounding as the intermediate reasoning stage, where the predicted bounding boxes serve as simple but effective \textit{visual chain-of-thought (CoT)} signals to help improve the quality of the final reasoning step.

The proposed method, \textbf{\ourwork}, is benchmarked across a diverse set of evaluation datasets, excelling not only in multimodal visual reasoning tasks~\cite{yue2024mmmu,eyeswideshut,vstar,cambrian1,grok,masry2022chartqa,textvqa,mathew2021docvqa,hudson2019gqa}, but also in object-centric visual grounding tasks~\cite{referitgame,refcoco}. We also systematically analyze different designs of the visual attention re-engagement mechanism, and its collaboration with involuntary attention in MLLMs. We hope that our study paves the way towards stronger multimodal intelligence by emphasizing more vision-centric and perception-driven reasoning mechanisms.

%% file: sections_ready/2_related.tex
\begin{figure}[!t]
    \centering
    \captionsetup{type=figure}
    \includegraphics[trim=0 0 0 0, clip=True, width=1\linewidth]{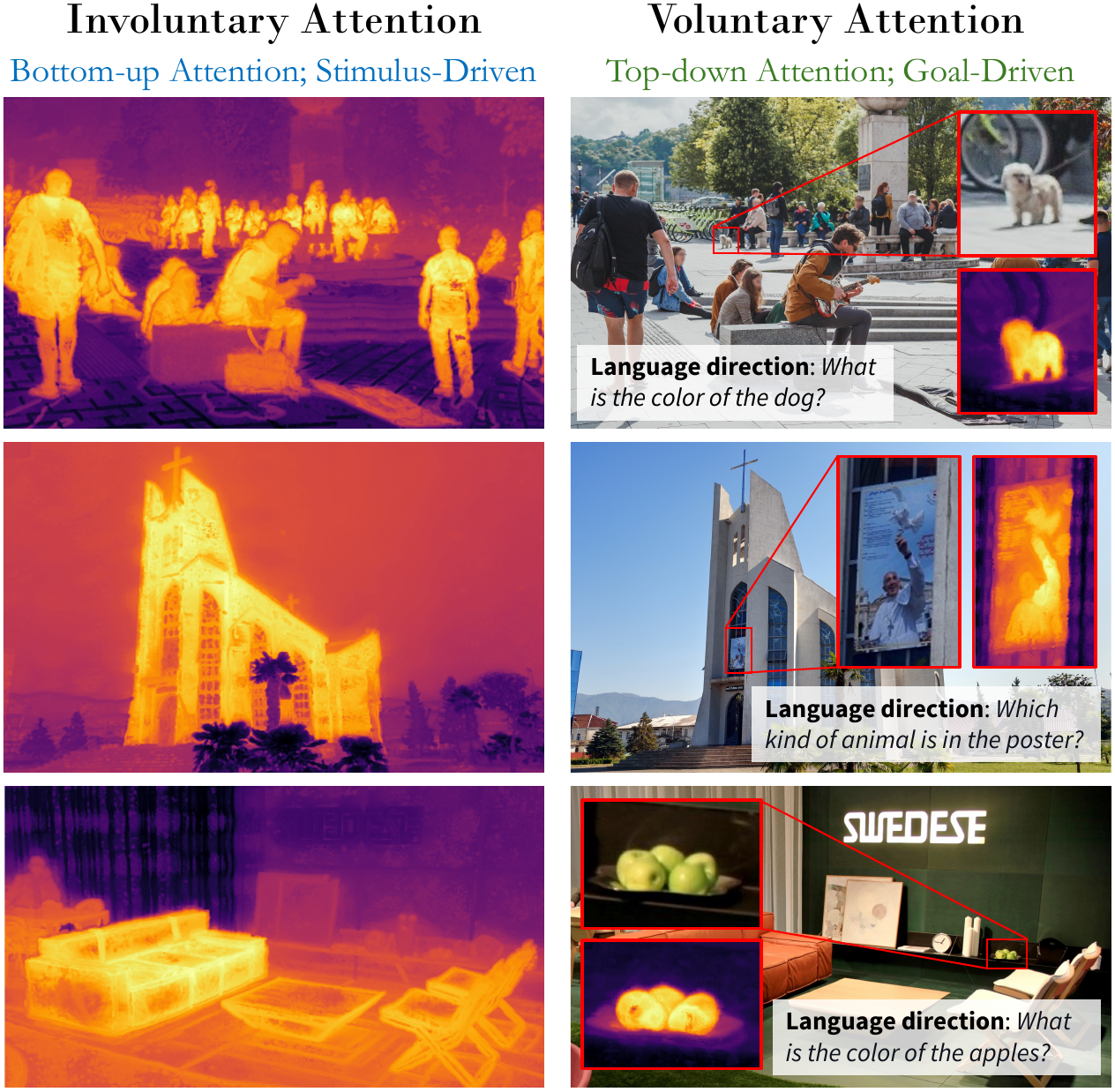}
    \vspace{-7mm}
    \captionof{figure}{Illustration of two visual attention mechanisms. \textit{Involuntary Attention (Left)}: stimulus-driven; unconditioned feature extraction; salient objects. \textit{Direct Attention (Right)}: Goal-driven; language-guided region-of-interest (RoI) feature extraction.
    } 
    \vspace{-4mm}
    \label{fig:two_attentions}
\end{figure}

\section{Related Work}
\label{sec:related_work}

\mypar{Visual Reasoning with MLLMs.} The emergence of multimodal large language models (MLLMs) has revolutionized visual reasoning capabilities, allowing sophisticated question-answering and complex visual understanding tasks. Visual instruction tuning~\cite{llava} pioneers this advancement by establishing a foundation for tuning language models to handle multimodal tasks effectively. Following this, several improvements and architectural innovations have emerged to enhance zero-shot generalization abilities of MLLMs, including better visual-linguistic alignment~\cite{vila,chen2024far,gao2024mini,Qwen-VL}, high-resolution visual input~\cite{llavanext,llavaonevision,Qwen2VL}, and dataset curation pipelines~\cite{chen2023internvl,cambrian1,Qwen2VL}. This progress extends to proprietary models like GPT-4V~\cite{gpt4v}, Claude 3~\cite{claude}, and Gemini~\cite{gemini,gemini15}, which have showcased remarkable general-purpose applicability.

Several recent studies have shifted focus towards exploring visual reasoning from a vision-centric perspective. Cambrian-1~\cite{cambrian1} conducts comprehensive investigations into various visual encoder architectures and introduces a specialized benchmark CV-Bench~\cite{cambrian1} for assessing vision-centric reasoning capabilities. Eagle~\cite{eagle} further advances this direction by introducing the mixture-of-vision-experts mechanism, demonstrating the potential of specialized visual processing highways in MLLM architectures. However, despite these advancements, current approaches lack conscious control over visual attention mechanisms and do not incorporate explicit goal-driven strategies for visual token extraction, which motivates the introduction of our method.

\vspace{2mm}
\mypar{Visual Perception with MLLMs.}
Visual perception has consistently been a critical and challenging task within the field of computer vision, encompassing fundamental tasks such as classification, detection, segmentation, and captioning. Numerous specialized models~\cite{radford2021clip,li2022blip, groundingdino, kirillov2023segany, ren2024grounded}, or ``vision experts," have been developed to tackle these tasks.
The emergence of MLLMs~\cite{llava, gpt4v} offers new opportunities to perception tasks. One line of work aims to construct multimodal agents that use MLLMs as controllers to activate specific visual experts~\cite{liu2023llavaplus, wu2023visual}. While this framework shows promising performance, it remains complex and unwieldy, limiting its practical applicability.
Another line of work involves building unified MLLMs to handle a wide range of vision tasks~\cite{ferret, xiao2024florence, wang2023cogvlm}. Although this approach is feasible in certain scenarios, its overall performance still falls short compared to specialized vision models.
Recently, more attention has shifted toward unified models that aspire to be comprehensive, spanning both understanding and generation tasks with the support of vast datasets~\cite{wang2024emu3, team2024chameleon}. 
Despite these advancements, most efforts have focused less on the synergy between visual perception and reasoning. In contrast, our research explores a grounded CoT approach to vision-centric reasoning, leveraging visual perception as a foundational component.

\vspace{2mm}
\mypar{Chain-of-thought Reasoning.} Chain-of-Thought (CoT) reasoning, first introduced by~\citet{cot}, demonstrates that prompting language models to generate intermediate reasoning steps significantly improves their problem-solving capabilities. This concept has sparked numerous works to further enhance reasoning performance, including zero-shot reasoning~\cite{zeroshotcot}, automatic CoT prompt generation~\cite{automaticcot}, and techniques like self-consistency prompting~\cite{selfconsistencycot}. Recent works have expanded beyond traditional linear reasoning paths, introducing more sophisticated frameworks such as Graph-of-Thoughts for complex problem decomposition~\cite{besta2024graphcot}, Program of Thoughts for structured numerical reasoning,~\cite{chen2023programcot} and Tree of Thoughts for deliberate decision-making~\cite{yao2024treecot}. In the multimodal domain, researchers have begun adapting CoT principles to vision-language tasks~\cite{lu2022learn2explain,ge2023multimodalcot,zhang2023multimodalcot,zheng2023ddcot,rose2023visualcotvqa,vocot}, using off-the-shelf object detectors or multi-turn visual instructions to improve visual reasoning for ambiguous instructions~\cite{vstar,ni2024visualo1,zhang2024improvevisualcot,shao2024visualcot}, or interleaved segmentation and question answering to develop joint perception and reasoning models~\cite{rasheed2024glamm,zhang2025psalm,zhang2024groundhog}. While these works demonstrate the potential of CoT in multimodal contexts, our work, \ourwork, is the first to systematically study the mechanism of explicit visual attention engagement as visual CoT signals within the MLLM design space, bridging the gap between grounding and vision-centric VQA tasks, while achieving state-of-the-art performance across diverse benchmarks in both domains.

%% file: sections_ready/3_method.tex
\section{Model}
\label{sec:method}

\begin{figure*}[!t] 
    \centering
    \captionsetup{type=figure}
    \includegraphics[trim=0 0 0 0, clip=True, width=1\linewidth]{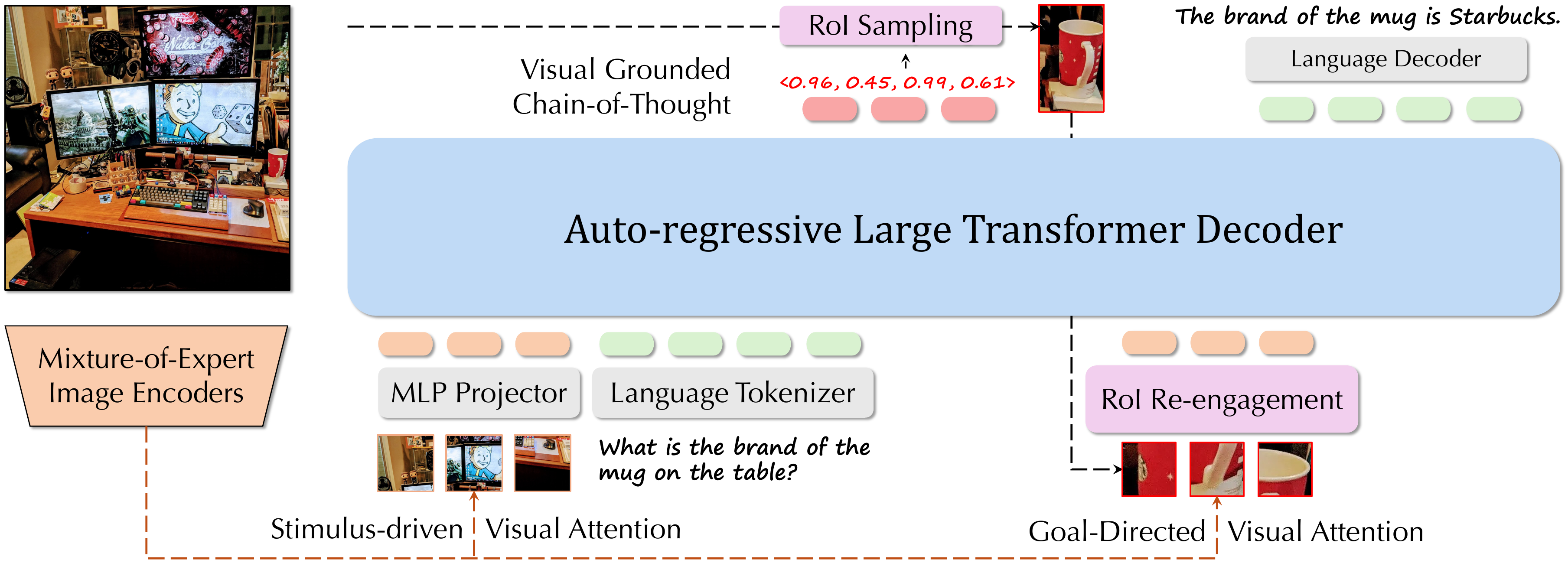}
    \vspace{-6mm}
    \captionof{figure}{Illustration of \ourwork architecture. In addition to standard unconditioned visual tokenization, our method incorporates an additional goal-directed visual tokenization procedure. The model has the ability to ground most relevant region-of-interest (RoI) conditioned on the multimodal input instructions. Then, the visual RoI is sampled from the input image, and fed to the RoI re-engagement module to extract another set of visual tokens as CoT context for reasoning.     
    } 
    \vspace{-4mm}
    \label{fig:architecture}
\end{figure*}

In this section, we demonstrate how we design the architecture of \ourwork from a vision-centric perspective. We start by revisiting the general design space of recent MLLMs~\cite{llava,eagle,llavaonevision,vila,cambrian1,minigemini,llavanext}. Most existing open-source visual reasoning MLLMs follow a unified autoregressive architectural paradigm, where input images are first converted into visual tokens and concatenated with language tokens before being jointly processed by the LLM for answer token prediction and decoding. This transformation process employs a visual encoder, typically a Vision Transformer (ViT-based~\cite{dosovitskiy2020vit}) architecture such as CLIP~\cite{radford2021clip}, followed by a multilayer perceptron (MLP-based) projector that maps visual features into the text token space. Although this unified architecture has demonstrated strong multimodal capabilities, \textit{it is not optimized for vision-centric scenarios that require precise visual focus for accurate reasoning}. To address this limitation, we propose \ourwork, a vision-centric reasoning framework with grounded chain-of-thought that enhances MLLMs through explicit language-directed visual region-of-interest (RoI) search and context re-engagement. We present our architectural designs in Section~\ref{sec:method:architectural-design}, and detail the directed visual context re-engagement module in Section~\ref{sec:method:visual-reengagement}. We validate our design choices and highlight the key empirical findings in Section~\ref{sec:analysis:ablation}. The complete \ourwork architecture is illustrated in Figure~\ref{fig:architecture}.

\subsection{Architectural Designs} 
\label{sec:method:architectural-design}

\mypar{Visual Encoders.} In multimodal vision-centric reasoning, visual encoders play a crucial role by ensuring minimal information loss during image-to-token abstraction and facilitating efficient vision-language alignment. We implement a mixture-of-vision-experts (MoVEs) strategy in our visual encoder suite, building upon recent MLLM studies~\cite{cambrian1,eagle} that demonstrate the complementary benefits of combining different vision foundation models. Our encoding system incorporates three vision experts: CLIP~\cite{radford2021clip}, ConvNeXt~\cite{liu2022convnet}, and EVA-02~\cite{fang2023eva01,eva02}. Following extraction, the 2D embeddings are interpolated to uniform spatial dimensions and concatenated along the channel dimension~\cite{lin2023sphinx,eagle}. For multimodal connectors, we employ MLP projectors, consistent with the practice of leading MLLM architectures~\cite{llavanext,llavaonevision,cambrian1,minigemini,eagle}.

The combined visual and language tokens form a multimodal input sequence that is processed by an autoregressive large transformer for next-token prediction. We leverage state-of-the-art pretrained LLMs~\cite{dubey2024llama3,chiang2023vicuna} as our transformer decoder, due to their robust zero-shot reasoning capabilities.

\vspace{1mm}
\mypar{Region-of-Interest Sampling.} To enable explicit visual search, we incorporate region-of-interest (RoI) prediction capabilities, allowing our model to output bounding boxes corresponding to regions referenced in the question prompt. This approach is approximately equivalent to the object grounding task, except in the visual reasoning task, our approach extends beyond well-defined objects to handle arbitrary regions relevant to visual reasoning. To maintain a simple design, we adopt the text encoding strategy for bounding box representation, where bounding boxes are normalized into $[0,1]$ range, and represented in text format ($[x_{\min}, y_{\min}, x_{\max}, y_{\max}]$)~\cite{shikra,ferret,ferretv2}. This approach eliminates the complexity of training additional box or mask decoding heads~\cite{groundingdino,groundingdino15,zhang2025psalm,zhang2024groundhog,zhang2024llavagrounding}. As illustrated in Figure~\ref{fig:architecture}, the predicted bounding box guides the cropping process of relevant RoIs from the input image for subsequent visual context re-engagement.

\subsection{Directed Visual Context Re-engagement}
\label{sec:method:visual-reengagement}

The model-predicted bounding boxes represent the visual context most relevant to the current reasoning objective. To leverage these regions-of-interest (RoIs) effectively, we seek to direct the model's attention towards these critical areas, thus enhancing focus on context pertinent to the language-defined objectives. Figure~\ref{fig:two_attentions} illustrates the denoised attention map~\cite{yang2024denoisingvit} of our CLIP encoder, showcasing how RoI-specific visual attendance accentuates the essential visual cues aligned with the goal. However, the optimal method for directing this attention remains an under-explored challenge. We identify and categorize four distinct strategies for guiding MLLMs to engage with sampled bounding boxes and incorporate these approaches within the \ourwork architecture for unified comparison and analysis.

\vspace{1mm}
\noindent\textit{Implicit Self-Attention.} Most existing MLLMs do not adopt explicit visual search or attending modules~\cite{llava,llavanext,eagle,cambrian1}. Instead, they rely on the intrinsic ability of LLMs to attend to visual context through the global self-attention layers. This implicit approach to RoI engagement adopts a minimalistic design, offering simplicity but limited control over specific attention to the bounding boxes. 

\vspace{1mm}
\noindent\textit{Implicit Box Guidance.} This strategy extends beyond basic self-attention by predicting bounding boxes as special tokens or text coordinates, without explicit visual RoI re-engagement. Although predominantly employed in perception tasks~\cite{ferret,ferretv2,shikra,lin2023sphinx}, this design can be extended to visual reasoning scenarios, where bounding box predictions serve as chain-of-thought signals, nudging self-attention implicitly towards the RoIs for reasoning purposes. By maintaining the CoT in text format, the shift in attention becomes subtler, merging visual and text-centric cues without explicitly emphasizing visual tokens.   

\begin{figure}[!t] 
    \centering
    \captionsetup{type=figure}
    \includegraphics[trim=0 0 0 0, clip=True, width=1\linewidth]{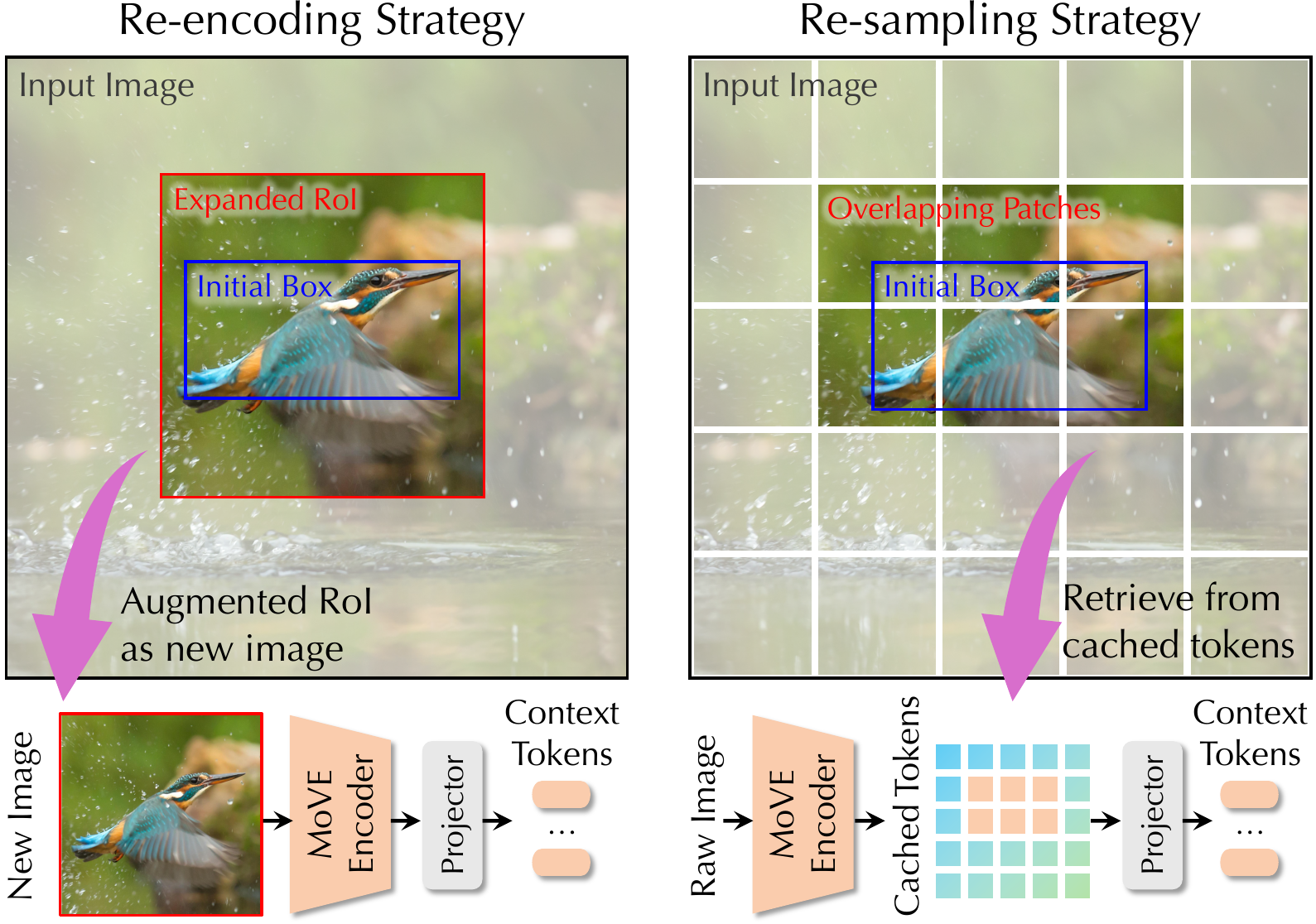}
    \vspace{-6mm}
    \captionof{figure}{Illustration of two visual CoT mechanisms. Re-encoding expand the RoI and treat it as a new image for tokenization. Re-sampling retrieves knowledge from the pre-extracted token cache.      
    } 
    \vspace{-4mm}
    \label{fig:reengagement}
\end{figure}

\vspace{1mm}
\noindent\textit{Explicit RoI Re-encoding.} In contrast to implicit methods, explicit RoI engagement represents visual CoT signals through actual visual tokens. As demonstrated in Figure~\ref{fig:reengagement} (\textit{left}), the re-encoding approach processes sampled image crops through vision encoders for tokenization~\cite{shao2024visualcot} after processing. The processing is equivalent to an augmentation process, which involves square padding of cropped regions to $\mathrm{max}\{\mathrm{width},\mathrm{height}\}$ region, context expansion with margins, and dimension-specific resizing for vision experts. These tokens are appended to the input sequence, introducing a supplementary visual context that guides reasoning through explicit, context-specific signals. This approach ensures that RoIs are attended to precisely, albeit with an increase in computational requirements due to the additional encoding process.

\vspace{1mm}
\noindent\textit{Explicit RoI Re-sampling.} The re-sampling method offers an alternative explicit engagement strategy that reduces computational overhead. As shown in Figure~\ref{fig:reengagement} (\textit{right}), rather than treating RoI boxes as new images, re-sampling utilizes visual embeddings from a memory bank~\cite{chang2021imageresample,kang2019decouplingresample,vocot}. In visual reasoning tasks, tokens are retrieved from the initial MoVE encoder suite and reused as needed. We calculate the overlap between the RoI bounding boxes and the patch embeddings after visual encoders, and resample the patch tokens that have intersection with boxes as context tokens for re-engagement. This strategy leverages cached tokens, thus streamlining computation while maintaining a focus on task-relevant regions. In the meanwhile, the redundant tokens also preserve the positional context within the original image, which might be lost during padding and resizing process in the re-encoding method.

%% file: sections_ready/4_analysis.tex
\input{tables/visual-reasoning-benchmarks}

\section{Experiments}
\label{sec:analysis}

This section presents our comprehensive experimental methodology and results. We begin by detailing our training protocols (Section~\ref{sec:analysis:training}), followed by implementation details (Section~\ref{sec:analysis:implementation}). We then describe our evaluation benchmarks and baseline comparisons (Section~\ref{sec:analysis:baselines}), and demonstrate the effectiveness of our model on vision reasoning and reference expression grounding tasks (Section~\ref{sec:analysis:main-performance}). Through extensive ablation studies, we validate the design choices of \ourwork in controlled settings (Section~\ref{sec:analysis:ablation}). More results and details are provided in the supplementary material.

\subsection{Training Pipeline}
\label{sec:analysis:training}

Following the advances of recent MLLMs~\cite{llava,llavanext,cambrian1,vila,eagle}, we divide the training into two stages. 

\vspace{1mm}
\mypar{Alignment and Pre-training.} In the initial pre-training stage, we adopt the LLaVA-595K dataset~\cite{llava}, which comprises carefully curated image-text pairs. We freeze the LLM while allowing the vision encoders and MLP projector layers to be trainable. Drawing insights from Eagle~\cite{eagle}, we implement a vision expert pre-alignment process to minimize representation disparities between experts and enhance the subsequent language alignment.

\vspace{1mm}
\mypar{Supervised Fine-Tuning (SFT).} The second stage employs a diverse combination of datasets to ensure robust performance across multiple domains. To ensure ensures strong general-purpose multimodal understanding capabilities, we utilize \textit{Eagle1.8M dataset}~\cite{eagle}, a comprehensive collection of conversational data aggregated from various sources~\cite{llava,mathew2021docvqa,kim2022ocrfree-syndogen,masry2022chartqa,kafle2018dvqa,kembhavi2016ai2d,chen2023sharegpt4v,laion-gpt4v,wang2023lvis-instruct4v,liu2023lrv-instruct,gao2023geollava,zhang2023llavar,zhu2016visual7w,openhermes2-5}. For visual chain-of-thought reasoning, we incorporate the \textit{VCoT dataset}~\cite{shao2024visualcot}, which provides region-of-interest (RoI) bounding box annotations specifically designed for grounding and reasoning tasks. This dataset obtains samples from multiple established benchmarks~\cite{textvqa,sidorov2020textcaps,mathew2021docvqa,van2023documentdude,huang2019sroie,wah2011caltechbirds,plummer2015flickr30k,zhu2016visual7w,mathew2022infographicvqa,liu2023vsr,hudson2019gqa,kuznetsova2020openimages}. We structure each sample as a multi-turn conversation between user and AI agent, where (1) The agent first predicts the region-of-interest using \texttt{\textless roi-box\textgreater} annotations in normalized text coordinates (Section~\ref{sec:method:architectural-design}). (2) The user then provides intermediate visual chain-of-thought signals through \texttt{\textless visual-context\textgreater} tokens. And (3) the agent generates the final response based on this structured interaction. To enhance our model's ability to ground concepts in unconstrained scenarios, we follow existing work~\cite{ferret,shikra,shao2024visualcot} by incorporating a mixture of GRIT~\cite{peng2023kosmos2} (756K) and Shikra~\cite{shikra,refcoco,krishna2017visualgenome,zhu2016visual7w,plummer2015flickr30k} (326K) datasets. All spatial grounding information is normalized to the range $[0, 1]$ relative to image dimensions and represented in text format. During this fine-tuning stage, we allow full parameter updates across the MoVE vision encoder, MLP projectors, and the LLM decoder.

\subsection{Implementation Details}
\label{sec:analysis:implementation}

We use Llama3-8B~\cite{dubey2024llama3} as our LLM decoder backbone. For vision encoders, we use ViT-L/14 CLIP~\cite{radford2021clip}, ConvNeXt-XXL-1024~\cite{liu2022convnet}, and EVA-02-L/16~\cite{eva02} as our MoVE encoding system. The input resolution is set to 1024$\times$1024 for ConvNeXt and EVA-02, and 448$\times$448 for CLIP model. The visual token number is 1024 (32$\times$32). Following Eagle~\cite{eagle}, we name our model \ourwork-X3, to reflect the usage of three vision experts. During the RoI selection, we encode RoI with the format of box coordinates: {\footnotesize $[x_{\min}, y_{\min}, x_{\max}, y_{\max}]$}, and the model is instruction-tuned to directly output them in text, where the numbers are normalized to $[0, 1]$ by image dimensions and accurate to three decimal places. We parse the coordinates by removing brackets and commas, then convert the numbers back to box coordinates for grounding and resampling. For both stages, we train for one epoch with a batch size of 256. The learning rate is set to 1$e$-3 for the pre-training stage, and 2$e$-5 for the SFT stage. The AdamW optimizer~\cite{loshchilov2017adamw} with zero weight decay and a cosine learning rate scheduler is used. Experiments are conducted using NVIDIA A100 GPUs. More details are provided in the supplementary material.

\subsection{Baseline Models and Benchmarks}
\label{sec:analysis:baselines}

We compare \ourwork with state-of-the-art MLLMs with roughly the same parameter size, including Mini-Gemini-HD~\cite{minigemini}, LLaVA-NeXT~\cite{llavanext}, VisCoT~\cite{shao2024visualcot}, QwenVL~\cite{Qwen-VL}, InternVL~\cite{chen2023internvl}, and Eagle~\cite{eagle}. We use open-sourced Eagle-X3-8B~\cite{eagle} with the same MoVE encoder structure as our baseline architecture. For reference, we also include performance metrics from proprietary models~\cite{gpt4v} or models trained with orders of magnitude more (or even undisclosed) data~\cite{Qwen2.5-VL,internvl2}. We use the official evaluation metrics provided by the benchmarks if available. Otherwise, we follow the same evaluation setting as the recent MLLMs~\cite{eagle} for a fair comparison. For grounding tasks, we benchmark against both specialist and generalist models including MAttNet~\cite{mattnet}, TransVG~\cite{transvg}, UNITER~\cite{uniter}, VILLA~\cite{villa}, UniTAB~\cite{unitab}, MDETR~\cite{mdetr}, G-DINO~\cite{groundingdino}, OFA-L~\cite{ofal}, Shikra~\cite{shikra}, Ferret~\cite{ferret}, MiniGPT-v2~\cite{chen2023minigptv2}, InternVL2~\cite{internvl2}, and Qwen-VL~\cite{Qwen-VL}. We report Acc@0.5 as the metrics for the referring expression grounding task. All model performances are obtained from official public reports.

\vspace{1mm}
\mypar{Benchmarks.} We report the performance of the methods on various vision-language benchmarks~\cite{vstar,cambrian1,eyeswideshut,grok,masry2022chartqa,kim2022ocrfree-syndogen,textvqa,mathew2021docvqa,yue2024mmmu,mm1,rasheed2024glamm,hudson2019gqa}. We follow prior work~\cite{eagle,cambrian1,shao2024visualcot} and combine their evaluation benchmarks to cover a wide spectrum of \textit{vision-centric multimodal} evaluation settings where the keys to correctly answering questions lie in accurate visual understanding.

\subsection{Main Performance}
\label{sec:analysis:main-performance}

Our experimental evaluation focuses on two primary capabilities: \textit{visual reasoning} and \textit{referring expression grounding}. These complementary tasks assess different aspects of our model's multimodal understanding: visual reasoning examines comprehensive multimodal comprehension, while referring expression grounding evaluates precise vision-text alignment through localization tasks.

\vspace{1mm}\mypar{Performance on Visual Reasoning.}
We conducted evaluations across three categories of MLLM benchmarks: \textit{General Multimodal Reasoning}, \textit{Text-centric Understanding}, and \textit{Vision-centric Perception}. Performance is shown in Table~\ref{tab:vqa}. We observe several findings from the results. (1) First, \ourwork achieve state-of-the-art performance among public MLLMs with a comparable parameter count and training scale. Notably, our approach even surpasses several proprietary MLLMs, demonstrating its exceptional multimodal reasoning capabilities. (2) Furthermore, we observe substantial improvements in both vision-centric and text understanding tasks. These tasks typically require precise attention to specific visual elements, such as objects or textual components within images, to generate accurate responses. Improvement in these areas highlights the effectiveness of our goal-conditioned visual search mechanism and enhanced visual attention engagement strategies.

\vspace{1mm}\mypar{Performance on Referring Grounding.}
To evaluate our model's object grounding capabilities, we utilized the RefCOCO, RefCOCO+, and RefCOCOg benchmarks~\cite{refcoco}. The results are shown in Table~\ref{tab:referring}. Our method achieves leading performance among comparable-scale generalist MLLMs, highlighting its effectiveness in combining general-purpose reasoning with precise visual grounding. Our performance is competitive against Grounding-DINO-L~\cite{groundingdino}, a specialist model trained on a larger grounding-specific dataset and optimized for detection tasks. These results demonstrate that \ourwork not only excels in high-level reasoning tasks, but also achieves exceptional performance in fundamental visual perception and localization tasks. 

\vspace{1mm}\mypar{Qualitative Results.} In Figure~\ref{fig:qualitative} we demonstrate some qualitative results of \ourwork on the visually grounded CoT task and the referring grounding task. Our method is capable of achieving complicated visual reasoning tasks with the help of the visually grounded CoT mechanism.

\subsection{Ablation Study and Analysis}
\label{sec:analysis:ablation}

The current landscape of the MLLM community presents great challenges for perfectly fair quantitative evaluation and comparison on benchmarks, due to variations in data scale, model sizes, and architectural choices. Hence, in this section, we focus on rigorous and controlled experiments to validate our architectural designs and describe our key findings. We employ an accelerated and unified training schedule across all ablation experiments for fair comparison.

\input{tables/referring}

\vspace{-4mm}
\subsubsection{Visual Attention Re-engagement Analysis.} 
\label{sec:analysis:ablation:reengagement}
\vspace{-2mm}

We conducted a systematic evaluation of different visual context re-engagement mechanisms, with results presented in Table~\ref{tab:re-engagement}. (1) First, \textbf{the incorporation of CoT reasoning consistently enhances performance across both visual and text-based reasoning tasks}. The introduction of the implicit bounding box CoT guidance yields substantial improvements over implicit attention mechanisms, with explicit CoT reasoning providing even greater performance gains. (2) When comparing re-encoding and re-sampling strategies, we find that \textbf{re-sampling generally demonstrates superior performance across most benchmarks}. This advantage can be attributed to the preservation of contextual positional information and the avoidance of distribution shifts that typically occur during region rescaling. However, \textit{this pattern shows an interesting exception in the V-Star benchmark}~\cite{vstar}, which emphasizes the perception of small objects in complex scenes. In this specific context, re-encoding proves more effective, as it processes regions with larger patches, thereby preserving more fine-grained details and minimizing information loss.

\vspace{1mm}
\noindent We make further discussion over the selection of re-encoding and re-sampling strategies. From a high-level perspective, re-sampling involves retrieving tokens from a token cache that was pre-extracted in the initial unconditioned phase. Conversely, re-encoding augments the cropped patch and produces a new embedding, leveraging higher resolution and enhanced contextual focus.

\vspace{1mm}\mypar{Impact of Visual Encoder Capacity.} We begin by examining how re-encoding and re-sampling strategies perform with visual encoders of varying capacities, as illustrated in Table~\ref{tab:re-engagement-explicit} ($\Delta$ fewer encoder). We specifically evaluate performance changes when MoVE is replaced with a single CLIP encoder. We have two key observations: (1) \textbf{Higher-capacity vision encoders consistently yield improved performance}, which aligns with our expectations that a robust feature extractor enhances subsequent perception and reasoning tasks. (2) \textbf{Re-encoding depends less on a high-quality initial feature extraction compared to re-sampling.} Since re-sampling simply retrieves tokens from the cache, it does not have the opportunity to update visual token quality. If detail is lost during the initial extraction, re-sampling struggles to recover it, whereas re-encoding can more effectively refine the visual representation.

\input{tables/reengagement}

\input{tables/reengagement-explicit}

\begin{figure*}[!t] 
    \centering
    \captionsetup{type=figure}
    \includegraphics[trim=0 0 0 0, clip=True, width=1\linewidth]{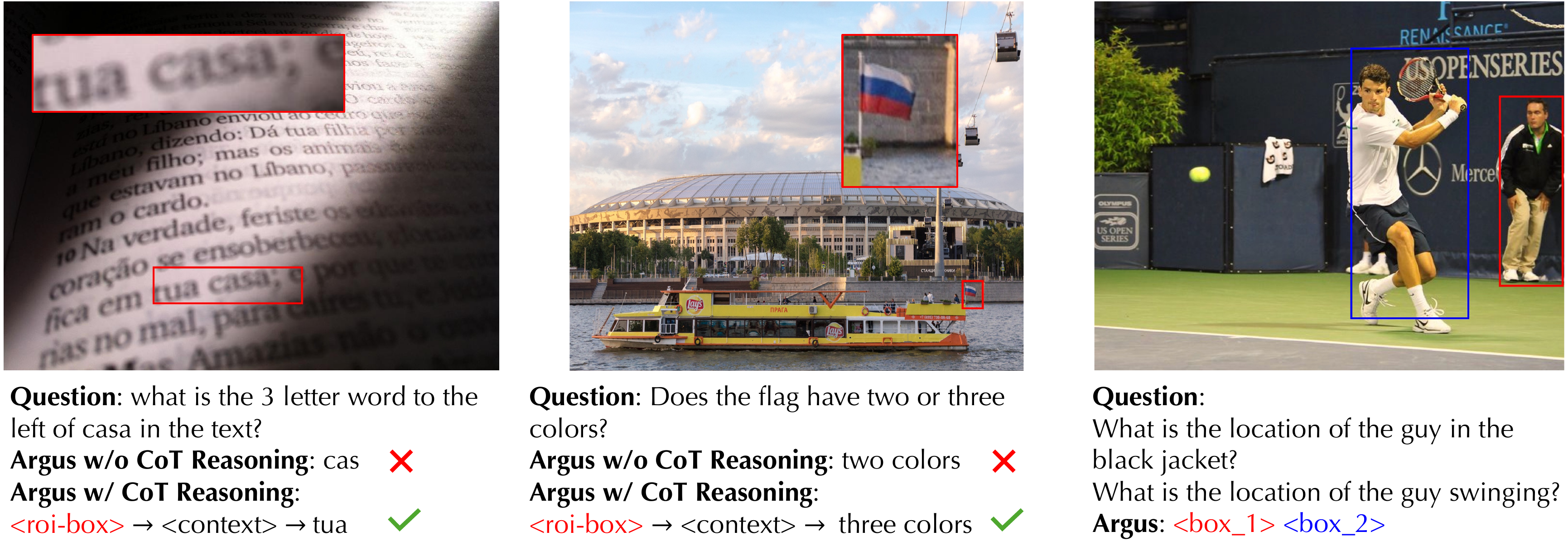}
    \vspace{-7mm}
    \captionof{figure}{Qualitative evaluation of \ourwork. We achieve superior performance in challenging multimodal reasoning and perception tasks.
    } 
    \vspace{-3mm}
    \label{fig:qualitative}
\end{figure*}

\vspace{1mm}\mypar{Effect of Padding and Square Context.} The model’s predicted RoI bounding boxes are typically rectangular, and preprocessing is required to convert them to square shapes, as most ViT-based vision encoders recommend. Table~\ref{tab:re-engagement-explicit} ($\Delta$ square context) shows how the re-encoding and re-sampling strategies respond to two different preprocessing methods. The default approach pads the image to make it square, while an alternative method expands the RoI to a square based on the larger dimension, effectively capturing more contextual information around the original focus area. Our results indicate that (1) \textbf{re-encoding consistently benefits from the larger context introduced by square bounding boxes}, while (2) re-sampling shows only marginal improvement in vision-centric reasoning and a noticeable drop in text-centric (OCR) reasoning. This outcome likely reflects the nature of text-centric tasks, which often involve locating specific words or sentences within chunks of text (as shown in Figure~\ref{fig:qualitative} (\textit{left})). In these cases, an expanded region may complicate localization tasks by introducing task-irrelevant context, reducing the benefits of bounding box prediction.

\vspace{1mm}\mypar{Non-shared MLP Layers.} For the re-sampling strategy, we also explore the use of non-shared MLP layers as an alternative to the default shared MLP configuration. This involves training a dedicated MLP layer specifically for the RoI re-engagement module. Results shown in Table~\ref{tab:re-engagement-explicit} ($\Delta$ non-share MLPs) suggest that separating MLP layers marginally improves performance. We attribute this improvement to the ability of the non-shared MLPs to account for distinct image distributions: one MLP is optimized for raw images with full context, while the other focuses on localized, object-centric regions. This approach effectively combines elements of both re-sampling and re-encoding, resulting in the best overall performance.

\input{tables/grounding-reasoning}

\vspace{-4mm}
\subsubsection{Grounding and Reasoning.} 
\vspace{-2mm}

We posit that visually grounded CoT directly connects the grounding and reasoning processes. In Table~\ref{tab:grounding-reasoning}, we analyze the impact of CoT signaling and grounding on reasoning performance, with two main observations: (1) By introducing the CoT dataset and re-engagement mechanism into SFT training, our method shows marked improvement in both vision- and text-centric reasoning tasks compared to the Eagle-X3, which is trained on a standard vision-language reasoning dataset, relying on implicit self-attention for attending to visual token. This highlights the importance of CoT for MLLMs. (2) Our full model, \ourwork, further incorporates grounding datasets into SFT training and exhibits additional performance gains. This enhancement can be attributed to its strengthened object-centric perception, which improves bounding box predictions and, in turn, maximizes the utility of the CoT mechanism.

\vspace{-4mm}
\subsubsection{Computational Efficiency.} 
\vspace{-2mm}

We compare the computational overheads introduced by re-encoding and re-sampling strategies. Table~\ref{tab:computation} shows that re-sampling has a major efficiency advantage in \textit{average visual encoding operations}, measured in giga-multiply-accumulate operations (GMACs), and \textit{number of additional visual tokens}, due to the reuse of patch embeddings, leading to a faster inference time, as LLM prediction is not blocked by visual encoding.

\input{tables/computation}

%% file: tables/visual-reasoning-benchmarks.tex
\begin{table*}[!t]
    \centering
    \fontsize{4pt}{4.8pt}\selectfont
    \setlength\tabcolsep{2pt} % Default value: 6pt
    \renewcommand{\arraystretch}{1.1} % Adjusts the row height
    \scalebox{2.21}{
    \begin{tabular}{r |cccccc |ccccc |ccccc}
     \multicolumn{1}{c}{Model} &
     \multicolumn{6}{c}{Vision-Centric Tasks} &
     \multicolumn{5}{c}{Text Understanding} &
     \multicolumn{5}{c}{General Tasks} \\
      
      Method &
      \rotatebox{90}{Avg} &
      \rotatebox{90}{V-Star} &
      \rotatebox{90}{CV-Bench$^\text{2D}$} &
      \rotatebox{90}{CV-Bench$^\text{3D}$} &
      \rotatebox{90}{MMVP} &
      \rotatebox{90}{RealworldQA} &

      \rotatebox{90}{Avg} &
      \rotatebox{90}{ChartQA} &
      \rotatebox{90}{OCRBench} &
      \rotatebox{90}{TextVQA}  &
      \rotatebox{90}{DocVQA}  &
      
      \rotatebox{90}{Avg} &
      \rotatebox{90}{MMMU$^\text{V}$} &
      \rotatebox{90}{MMB} &
      \rotatebox{90}{SEED$^\text{I}$} &
      \rotatebox{90}{GQA}
      \\
      \hline
      
      \textcolor{gray!130}{\textit{ref}: GPT-4o (\textit{250128})~\cite{gpt4v}}
      & \textcolor{gray!130}{73.7} & \textcolor{gray!130}{70.7} & \textcolor{gray!130}{79.8} & \textcolor{gray!130}{84.3} & \textcolor{gray!130}{58.5} & \textcolor{gray!130}{75.4} & \textcolor{gray!130}{83.1} & \textcolor{gray!130}{86.9} & \textcolor{gray!130}{75.1} & \textcolor{gray!130}{79.7} & \textcolor{gray!130}{90.8} & \textcolor{gray!130}{-} & \textcolor{gray!130}{68.9} & \textcolor{gray!130}{87.1} & \textcolor{gray!130}{81.7} & \textcolor{gray!130}{-}
      \\
      
      \textcolor{gray!130}{\textit{ref}: Qwen2.5-VL~\cite{Qwen2.5-VL}} 
      & \textcolor{gray!130}{72.6} & \textcolor{gray!130}{72.8} & \textcolor{gray!130}{80.0} & \textcolor{gray!130}{82.7} & \textcolor{gray!130}{53.1} & \textcolor{gray!130}{74.3} & \textcolor{gray!130}{87.9} & \textcolor{gray!130}{85.2} & \textcolor{gray!130}{85.9} & \textcolor{gray!130}{84.9} & \textcolor{gray!130}{95.7} & \textcolor{gray!130}{-} & \textcolor{gray!130}{58.6} & \textcolor{gray!130}{86.5} & \textcolor{gray!130}{78.3} & \textcolor{gray!130}{-}
      \\
      
      \textcolor{gray!130}{\textit{ref}: InternVL2.5~\cite{internvl2}}
      & \textcolor{gray!130}{71.1} & \textcolor{gray!130}{69.1} & \textcolor{gray!130}{79.7} & \textcolor{gray!130}{81.5} & \textcolor{gray!130}{54.9} & \textcolor{gray!130}{70.1} & \textcolor{gray!130}{84.8} & \textcolor{gray!130}{84.8} & \textcolor{gray!130}{82.2} & \textcolor{gray!130}{79.1} & \textcolor{gray!130}{93.0} & \textcolor{gray!130}{-} & \textcolor{gray!130}{56.0} & \textcolor{gray!130}{84.6} & \textcolor{gray!130}{79.2} & \textcolor{gray!130}{-}
      \\
                 
      \hline
      \rowcolor{gray!20}
      \multicolumn{17}{l}{\textit{7B \& 8B Open MLLMs, and Visual-CoT Variants}} 
      \\

      Vanilla CoT Prompting
      & 60.3 & 64.9 & 66.1 & 63.5 & 44.2 & 62.8 & 68.2 & 70.9 & 56.7 & 70.8 & 74.2 & 62.5 & 40.3 & 69.7 & 75.0 & 64.8
      \\

      Visual-CoT-7B~\cite{shao2024visualcot} 
      & 54.4 & 49.7 & 61.5 & 62.4 & 35.7 & 62.9 & 66.6 & 69.7 & 51.6 & 70.0 & 75.1 & 60.2 & 37.2 & 67.3 & 74.1 & 62.0
      \\

      Mini-Gemini-HD-8B~\cite{minigemini} 
      & 51.8 & 52.9 & 62.2 & 63.0 & 18.7 & 62.1 & 62.9 & 59.1 & 47.7 & 70.2 & 74.6 & 61.9 & 37.3 & 72.7 & 73.2 & 64.5
      \\
      
      LLaVA-NeXT-8B~\cite{llavanext} 
      & 55.4 & 50.8 & 62.2 & {65.3} & 38.7 & 60.1 & 63.9 & 69.5 & 49.0 & 64.6 & 72.6 & 62.9 & 41.7 & 72.1 & 72.7 & {65.2}
      \\

      InternVL~\cite{chen2023internvl}
      & 52.5 & 52.4 & 59.2 & 54.6 & 36.0 & 60.4 & 62.5 & 70.1 & 51.1 & 61.5 & 67.3 & 57.6 & 35.3 & 64.1 & 66.9 & 63.9
      \\

      QwenVL-7B~\cite{Qwen-VL}
      & 52.0 & 54.5 & 57.5 & 55.9 & 33.3 & 58.8 & 60.9 & 66.3 & 50.6 & 61.5 & 65.1 & 56.5 & 35.9 & 67.0 & 65.4 & 57.5
      \\

      Eagle-X3-8B~\cite{eagle} 
      & 59.6 & 60.7 & 66.4 & 63.0 & 45.1 & 62.9 & 67.8 & 70.4 & 56.1 & 70.9 & 73.9 & 62.4 & 39.8 & 70.9 & 73.9 & 64.9
      \\

      \hline
      \ourwork-X3-8B (ours)
      & \textbf{65.3} & \textbf{78.5} & \textbf{68.5} & \textbf{69.6} & \textbf{45.5} & \textbf{64.6} & \textbf{70.1} & \textbf{74.8} & \textbf{56.7} & \textbf{73.6} & \textbf{75.4} & \textbf{63.4} & \underline{40.4} & \textbf{72.9} & \textbf{75.8} & \underline{65.1}
      \\
   \hline  
   \end{tabular}
}
\vspace{-3mm}
\caption{{\ourwork achieve state-of-the-art performance among public MLLMs of comparable parameter count and training scale.}}
\vspace{-4mm}
\label{tab:vqa}
\end{table*}

%% file: tables/referring.tex
\begin{table}[t]
    \centering
    \fontsize{4pt}{4.8pt}\selectfont
    \setlength\tabcolsep{2pt} % Default value: 6pt
    \renewcommand{\arraystretch}{1.1} % Adjusts the row height
    \scalebox{1.81}{
    \begin{tabular}{r |ccc|ccc|cc}
     \multicolumn{1}{c}{Model} &
     \multicolumn{3}{c}{RefCOCO} &
     \multicolumn{3}{c}{RefCOCO+} &
     \multicolumn{2}{c}{RefCOCOg} \\
       &
      \rotatebox{0}{val} &
      \rotatebox{0}{testA} &
      \rotatebox{0}{testB} &
      \rotatebox{0}{val} &
      \rotatebox{0}{testA} &
      \rotatebox{0}{testB} &
      \rotatebox{0}{val} &
      \rotatebox{0}{test} \\
      
      \hline
      \rowcolor{gray!20}
      \multicolumn{9}{l}{\textit{Specialist Models}}
      \\

      MAttNet\cite{mattnet} & 76.4 & 80.4 & 69.3 & 64.9 & 70.3 & 56.0 & 66.7 & 67.0\\
      
      TransVG~\cite{transvg} & 81.0 & 82.7 & 78.3 & 64.8 & 70.7 & 56.9 & 68.7 & 67.7\\
      
      UNITER~\cite{uniter} & 81.4 & 87.0 & 74.2 & 75.9 & 81.5 & 66.7 & 74.0 & 68.7\\

      VILLA~\cite{villa} & 82.4 & 87.5 & 74.8 & 76.2 & 81.5 & 66.8 & 76.2 & 76.7 \\

      UniTAB~\cite{unitab} & 86.3 & 88.8 & 80.6 & 78.7 & 83.2 & 69.5 & 80.0 & 80.0\\
      
      MDETR~\cite{mdetr} & 86.8 & 89.6 & 81.4 & 79.5 & 84.1 & 70.6 & 81.6 & 80.9\\

      G-DINO~\cite{groundingdino} & 90.6 & 93.2 & 88.2 & 82.8 & 89.0 & 75.9 & 86.1 & 87.0 \\

      \hline
      \rowcolor{gray!20}
      \multicolumn{9}{l}{\textit{Generalist MLLMs}}
      \\

      OFA-L~\cite{ofal} & 80.0 & 83.7 & 76.4 & 68.3 & 76.0 & 61.8 & 67.6 & 67.6\\

      Shikra-7B~\cite{shikra} & 87.0 & 90.6 & 80.2 & 81.6 & 87.4 & 72.1 & 82.2 & 82.2\\

      Ferret-7B~\cite{ferret} & 87.5 & 91.4 & 82.5 & 80.8 & 87.4 & 73.1 & 83.9 & 84.8\\

      MiniGPT-7B~\cite{chen2023minigptv2} & 88.7 & 91.7 & 85.3 & 80.0 & 85.1 & 74.5 & 84.4 & 84.7 \\

      InternVL2-8B~\cite{internvl2} & 87.1 & 91.1 & 80.7 & 79.8 & 87.9 & 71.4 & 82.7 & 82.7 \\

      QwenVL-7B~\cite{Qwen-VL} & 89.4 & 92.3 & 85.3 & 83.1 & 88.3 & {77.2} & 85.6 & {85.5} \\
      
      \textbf{\ourwork-X3-8B} & \textbf{89.8} & \textbf{92.9} & \textbf{85.4} & \textbf{84.7} & \textbf{90.1} & \underline{77.1} & \textbf{86.7} & \underline{85.2}
      \\

   \hline  
   \end{tabular}
}
\vspace{-2mm} 
\caption{{Our method achieves leading performance among generalist MLLMs of comparable scale, highlighting its effectiveness in general-purpose reasoning with precise visual grounding.}}
\vspace{-4mm}
\label{tab:referring}
\end{table}

%% file: tables/reengagement.tex
\begin{table}[!t]
    \centering
    \fontsize{4pt}{4.8pt}\selectfont
    \setlength\tabcolsep{2pt} % Default value: 6pt
    \renewcommand{\arraystretch}{1.1} % Adjusts the row height
    \scalebox{2.15}{
    \begin{tabular}{r |cccc}
      Method &
      \rotatebox{0}{V-Star} &
      \rotatebox{0}{CV-Bench-{2D}}  &
      \rotatebox{0}{TextVQA} &
      \rotatebox{0}{ChartQA} \\
      \hline
      Implicit Att. & 58.6 & 64.5 & 69.2 & 67.3  \\
      Box Guidance. & 63.9 & 67.0 & \underline{71.6} & 70.4   \\
      RoI Re-enc. & \textbf{68.1} & \underline{67.4} & 71.4 & \underline{71.8}   \\
      RoI Re-smp. & \underline{67.0} & \textbf{68.2} & \textbf{73.9} & \textbf{72.7}   \\
   \hline  
   \end{tabular}
}
\vspace{-2mm}
\caption{Comparison of four visual attention re-engagement mechanisms, as introduced in Section~\ref{sec:method:visual-reengagement}, respectively. Two explicit visual RoI re-engagement mechanism leads the performance.}
\vspace{-4mm}
\label{tab:re-engagement}
\end{table}

%% file: tables/reengagement-explicit.tex
\begin{table}[!t]
    \centering
    \fontsize{4pt}{4.8pt}\selectfont
    \setlength\tabcolsep{2pt} % Default value: 6pt
    \renewcommand{\arraystretch}{1.1} % Adjusts the row height
    \scalebox{1.99}{
    \begin{tabular}{l |cccc}
      Method &
      \rotatebox{0}{V-Star} &
      \rotatebox{0}{CV-Bench-{2D}}  &
      \rotatebox{0}{TextVQA} &
      \rotatebox{0}{ChartQA} \\
      \hline
      
      \rowcolor{gray!20}
      \multicolumn{1}{l|}{{Re-encoding}} &  \multicolumn{4}{l}{}
    \\
      $\Delta$ fewer encoder & \textcolor{red}{--4.2} & \textcolor{red}{--3.8} & \textcolor{red}{--4.7} & \textcolor{red}{--3.9}  \\
      $\Delta$ square context & \textcolor{mygreen}{+1.6} & \textcolor{mygreen}{+0.8} & \textcolor{mygreen}{+0.1} & \textcolor{mygreen}{+2.6}   \\
      
      \hline
      \rowcolor{gray!20}
      \multicolumn{1}{l|}{{Re-sampling}} &  \multicolumn{4}{l}{}
    \\
      $\Delta$ fewer encoder & \textcolor{red}{--11.5} & \textcolor{red}{--6.1} & \textcolor{red}{--4.6} & \textcolor{red}{--5.3}  \\
      $\Delta$ square context & \textcolor{mygreen}{+0.5} & \textcolor{mygreen}{+0.1} & \textcolor{red}{--3.1} & \textcolor{red}{1.9}   \\
      $\Delta$ non-share MLPs & {0} & \textcolor{mygreen}{+1.7} & \textcolor{mygreen}{+0.5} & \textcolor{mygreen}{+0.3}   \\
      
   \hline  
   \end{tabular}
}
\vspace{-2mm}
\caption{Performance change ($\Delta$) after changing design choices between two types of explicit visual re-engagemment strategies. We use red for decrease $\downarrow$ and green for increase $\uparrow$.}
\vspace{-4mm}
\label{tab:re-engagement-explicit}
\end{table}

%% file: tables/grounding-reasoning.tex
\begin{table}[!t]
    \centering
    \fontsize{4pt}{4.8pt}\selectfont
    \setlength\tabcolsep{2pt} % Default value: 6pt
    \renewcommand{\arraystretch}{1.1} % Adjusts the row height
    \scalebox{2.05}{
    \begin{tabular}{r |cccc}
      Method &
      \rotatebox{0}{V-Star} &
      \rotatebox{0}{CVB-{2D}}  &
      \rotatebox{0}{TextVQA} &
      \rotatebox{0}{ChartQA} \\
      \hline
      \textit{Baseline} (Eagle-X3)         & 55.3 & 64.9 & 66.3 & 63.0  \\
      + \textit{CoT signals}          & 62.7 & 65.5 & 71.1 & 69.4   \\
      ++ \textit{Grounding} (\ourwork)    & 67.0 & 68.2 & 73.9 & 72.7   \\
   \hline  
   \end{tabular}
}
\vspace{-2mm}
\caption{Impact of CoT and grounding on reasoning performance. We verify that the combination of CoT mechanism and grounding task both benefits the model reasoning capability.}
\vspace{-4mm}
\label{tab:grounding-reasoning}
\end{table}

%% file: tables/computation.tex
\begin{table}[!t]
    \centering
    \fontsize{4pt}{4.8pt}\selectfont
    \setlength\tabcolsep{2pt} % Default value: 6pt
    \renewcommand{\arraystretch}{1.1} % Adjusts the row height
    \scalebox{2.55}{
        \begin{tabular} {l|ccc} 
        Modules 
        & \texttt{GMACs} & visual tokens & time$^{\text{inference}}$
        \\\hline
        Re-encoding 
        & 8,710.6 & 1024 & 827 ms \\
        Re-sampling
        & \textbf{4,355.3} & \textbf{26} & \textbf{492} ms\\
        \hline
        \end{tabular}
        }
\vspace{-2mm}
\caption{The re-sampling strategy shows advantages in computational efficiency over the re-encoding strategy.}
\vspace{-4mm}
\label{tab:computation}
\end{table}

%% file: sections_ready/5_conclusion.tex
\section{Conclusion}
\label{sec:conclusions}

This work introduces \ourwork, a vision-centric reasoning model with grounded chain-of-thought capability. By incorporating a grounding-driven visual attention re-engagement mechanism, \ourwork demonstrates an effective approach to enhancing multimodal reasoning by emphasizing directed visual focus. Through extensive evaluations, we show that our framework demonstrates superior performance across both multimodal reasoning and referral object grounding tasks. These findings not only advance our understanding of vision-language fusion, but also suggest a promising direction for future MLLM architectures that emphasize vision-centric mechanisms and visual chain-of-thought as a crucial component of multimodal intelligence.

\paragraph{Acknowledgments.} We thank Subhashree Radhakrishnan, Fuxiao Liu, and Min Shi for the fruitful discussion on the idea of the project and the implementation of the codebase. Y. Man is supported by the NVIDIA Graduate Fellowship. Y. Man and Y.-X. Wang are supported in part by NSF Grant 2106825 and NIFA Award 2020-67021-32799.

%% file: sections_ready/X_supplementary.tex
\clearpage
\setcounter{page}{1}
\maketitlesupplementary

\renewcommand{\thesection}{\Alph{section}}
\renewcommand{\thefigure}{\Alph{figure}}
\renewcommand{\thetable}{\Alph{table}}
\setcounter{section}{0}
\setcounter{figure}{0}
\setcounter{table}{0}

% ========= Additional Experiments ==========

\section{Additional Experiments and Analysis}
\label{sec:supp:more_results}

\input{tables/supp_roi_expansion}

\subsection{RoI Context Expansion}

\begin{figure}[!t] 
    \centering
    \captionsetup{type=figure}
    \includegraphics[trim=0 0 0 0, clip=True, width=1\linewidth]{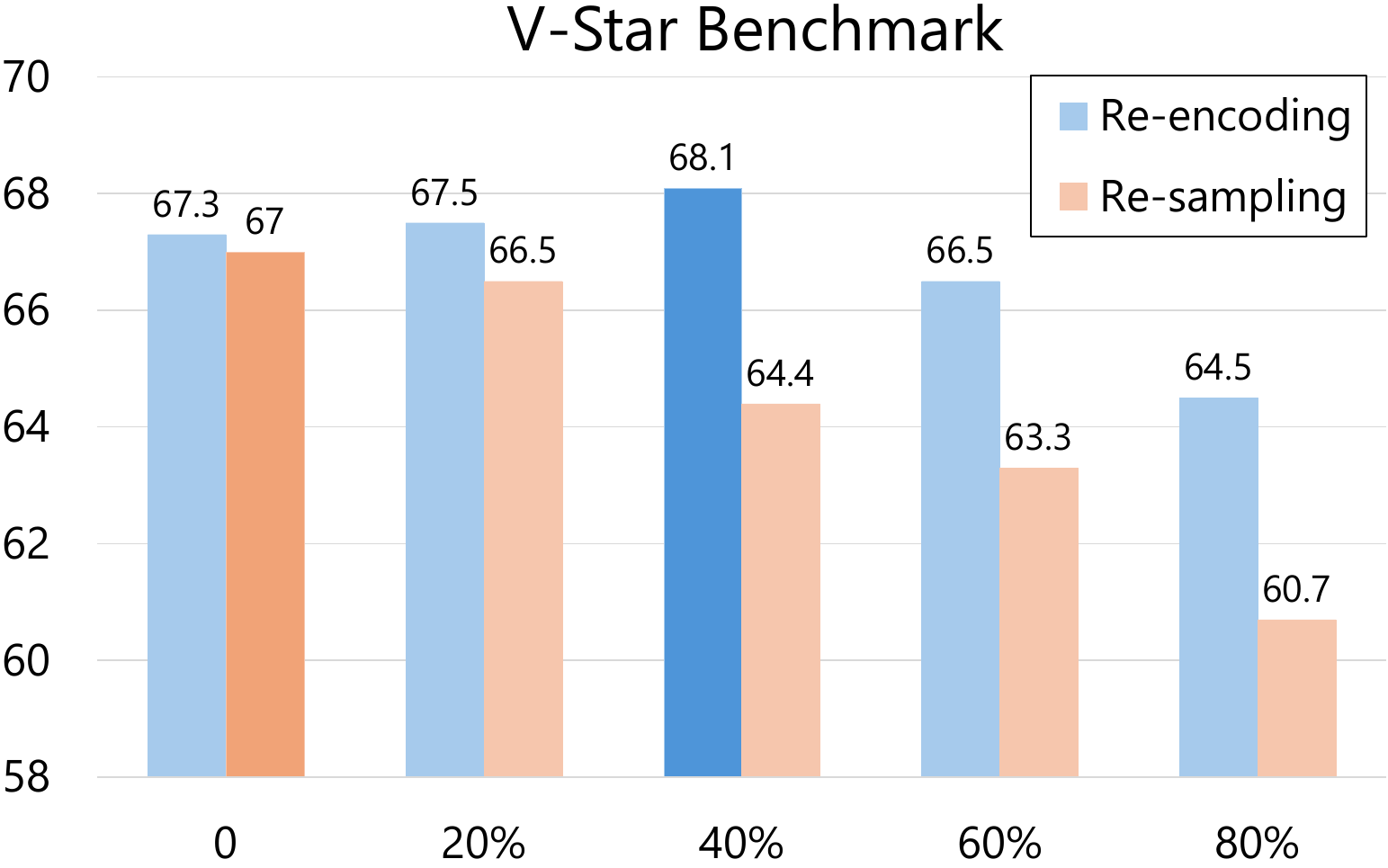}
    \includegraphics[trim=0 0 0 -30, clip=True, width=1\linewidth]{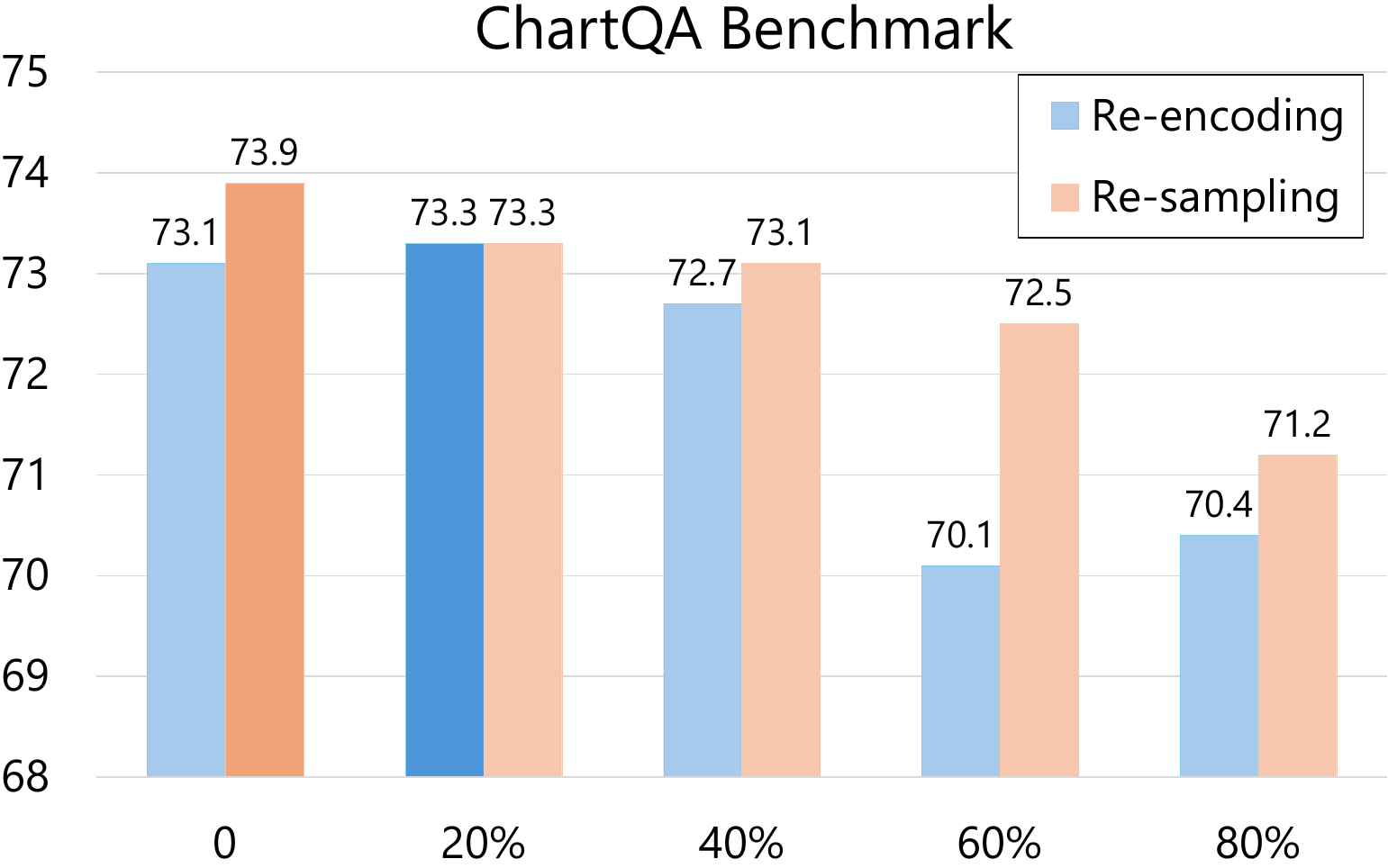}
    \vspace{-5mm}
    \captionof{figure}{Performance comparison of re-encoding and re-sampling strategies under varying region context expansion ratios. Re-encoding achieves optimal performance with an expanded context region (20\% to 40\% expansion), while re-sampling performs best with the original box size (0\% expansion). The optimal performance points for each strategy are highlighted in darker colors.
    } 
    \vspace{-4mm}
    \label{fig:supp:roi-context-expansion}
\end{figure}

To investigate the impact of region of interest (RoI) context expansion on \ourwork, we examined how expanding a predicted bounding box affects performance. Specifically, we expanded the bounding box by a fixed ratio to include additional context around the predicted center. If the expanded region exceeded the image boundaries, it was cropped to fit within them. Table~\ref{tab:supp:roi-context-expansion} and Figure~\ref{fig:supp:roi-context-expansion} present the performance evaluation of various expansion ratios using two distinct visual re-engagement strategies. Our results reveal the following insights.

\vspace{1mm}\mypar{Re-encoding Strategy.} The re-encoding approach benefits from a moderate expansion of the context region. Optimal performance is achieved with a 20 to 40\% expansion ratio on the ChartQA~\cite{masry2022chartqa} and V-Star~\cite{vstar} benchmarks. The additional context helps mitigate issues stemming from overly tight or slightly inaccurate bounding boxes, which are common in object grounding tasks. Moreover, the larger context aids in localizing the bounding box's relative position within the image, which is particularly beneficial for tasks that require both local and global context reasoning.

\vspace{1mm}\mypar{Re-sampling Strategy.} Unlike re-encoding, the re-sampling strategy method achieves its best performance with the original bounding box size. This can be attributed to an inherent context-expansion mechanism that leverages overlapping patches, utilizing all patch embeddings that intersect with the bounding box region as the input to the re-engagement module. As a result, further expansion of the bounding box does not yield additional benefits.

\vspace{1mm}\mypar{Effect of Excessive Expansion.} For both strategies, overly large context regions hurt performance. Including excessive irrelevant information introduces noise, which distracts the model from focusing on the most relevant RoIs. This counteracts the advantages of RoI grounding and diminishes overall effectiveness.

\input{tables/supp_LLM_variation}

\vspace{1mm}\mypar{Choice of LLMs and Architectural Designs.} In Table~\ref{tab:supp:different_llms}, we include Vicuna 8B, 13B, and Llama 8B as LLM backbones on Argus for comparison. The results show that our method generalizes to different LLMs. In addition, stronger LLMs (in size and data) lead to direct gains in performance.

\input{tables/supp_multi_roi}

\vspace{1mm}\mypar{Multi-RoI Scenarios.} The CoT reasoning instruction tuning datasets that we used in our work generally follow a single-RoI setting. In Table~\ref{tab:supp:multi_roi}, we extend \ourwork to multi-RoI reasoning with a simple multi-step framework. First, we (1) prompt the model to output RoIs (objects) about the questions in the text format. Then (2) after parsing, we conduct CoT reasoning for each separate object, and finally (3) merge multiple CoT grounding boxes and textual thoughts into one joint CoT signal for question answering. The results in the table show that this extension \textit{improves the performance by large margins} on two visual reasoning benchmarks~\cite{vstar,cambrian1}, demonstrating the flexibility of Argus in handling multi-RoI settings.

\vspace{1mm}\mypar{Discussion about Performance Discrepancy.} The performance discrepancy in CV-Bench$^{\text{3D}}$ is likely due to the bias of the multi-RoI data, which is not extensively covered in our training data. As a result, the results of the multi-RoI extension experiment shown in Table~\ref{tab:supp:multi_roi} demonstrate significant performance increase. For the performance discrepancy of MMMU~\cite{yue2024mmmu} and GQA~\cite{hudson2019gqa}, it is caused by strong language biases, as explained in~\cite{cambrian1} (Sec.~3.1). These two benchmarks depend more on language cues rather than visual input to correctly answer the questions, and thus we believe that a more language-oriented training data curation can lead to better performance.

\section{Additional Experiment Details} \label{sec:supp:more_implementation_details}

\subsection{Visual Foundation Models and LLMs}
\label{sec:supp:more_implementation_details:models}

\mypar{CLIP}~\cite{radford2021clip}. CLIP learns a unified embedding space for visual and textual content through contrastive learning. It optimizes the alignment between matching image-caption pairs while simultaneously pushing apart non-matching pairs in the embedding space. Due to its robust cross-modal understanding capabilities, it has established itself as the predominant vision encoder for multimodal large language models (MLLMs). In our implementation, we leverage the official huggingface checkpoint\footnote{\url{https://huggingface.co/openai/clip-vit-large-patch14-336}} of the ViT-L/14 architecture to initialize our CLIP vision expert. Following~\cite{eagle}, we interpolate the positional embedding to obtain the input image dimension $448\times 448$.

\vspace{1mm}
\mypar{ConvNeXt}~\cite{liu2022convnet}. ConvNeXt represents a modern evolution of convolutional neural networks (CNNs) that bridges the gap between CNNs and transformers. By incorporating transformer-inspired design principles while preserving the inherent advantages of convolutional architectures, it achieves exceptional performance across diverse vision tasks, making it an excellent choice as a vision expert. We employ the official checkpoint\footnote{\url{https://huggingface.co/timm/convnext_xxlarge.clip_laion2b_soup_ft_in1k}} of a ConvNeXt-XXLarge model, which has been pre-trained on LAION-2B and fine-tuned on ImageNet-1K. The input image dimension is set to $1024\times 1024$.

\vspace{1mm}
\mypar{EVA-02}~\cite{eva02,fang2023eva01}. EVA-02 is a vision foundation model that achieves superior performance with moderate model sizes. It incorporates Transformer architecture designs and utilizes masked image modeling pre-training with features from a large CLIP vision encoder. For this work, we specifically employ the EVA-02-L/16 model checkpoint pre-trained on detection-focused datasets including COCO~\cite{mscoco} and Objects365~\cite{shao2019objects365}, making it particularly well-suited for perceptive tasks. We utilize the official checkpoint\footnote{\url{https://huggingface.co/Yuxin-CV/EVA-02/blob/main/eva02/det/eva02_L_coco_det_sys_o365.pth}} and process input images at a resolution of $1024\times 1024$.

\vspace{1mm}
\mypar{Llama 3}~\cite{dubey2024llama3}. Llama 3 represents the latest advancement in open-sourced large language models (LLMs), incorporating significant improvements over its predecessors in instruction-following capabilities and reasoning ability. For our implementation, we employ the official checkpoint\footnote{\url{https://huggingface.co/meta-llama/Meta-Llama-3-8B-Instruct}} of the Meta-Llama-3-8B-Instruct model, which has been specifically fine-tuned for instruction-following scenarios.

\subsection{Implementation Details}

\mypar{Global Training Hyperparameters.} Table~\ref{tab:supp:hyper-global} details the global training hyperparameters we have employed across both stages of \ourwork training. For stage 1, we initialize the vision experts using pre-aligned checkpoints as described in Eagle~\cite{eagle}, while the MLP projector is randomly initialized. In stage 2, both the vision experts and MLP projectors are initialized using the checkpoints obtained from stage 1. The pre-training stage is trained with $32\times$ NVIDIA A100 GPUs for 4 hours, , while the supervised fine-tuning (SFT) stage utilizes $64\times$ NVIDIA A100 GPUs and requires 28 hours of training.

\begin{table}[!t]
\centering
  \begin{tabular}{l|cc}
    \toprule
    {Parameters} & Stage 1 & Stage 2\\
    \midrule
    Learning rate & $1\mathrm{e}^{-3}$ & $2\mathrm{e}^{-5}$\\
    Vision encoders & trainable & trainable \\
    Projector & trainable & trainable \\
    LLM backbone & frozen & trainable \\
    Global batch size & $256$ & $256$\\
    Optimizer & AdamW & AdamW\\
    Weight decay    & $0.0$ & $0.0$\\
    Beta coefficient $\beta_1$ & $0.9$ & $0.9$\\
    Beta coefficient $\beta_2$ & $0.999$ & $0.999$\\
    Epsilon coefficient $\epsilon$ & $1\mathrm{e}^{-8}$ & $1\mathrm{e}^{-8}$\\
    Gradient accumulation steps & $1$ & $1$\\
    Warmup ratio & $0.03$ & $0.03$\\
    Epochs     & $1$ & $1$\\
    Projector type & mlp$2\times$ & mlp$2\times$\\
    Learning rate scheduler & cosine & cosine\\
    Gradient checkpointing & true & true\\
    Precision & bfloat16 & bfloat16\\
    Max sequence length & 2048 & 3072\\
  \bottomrule
\end{tabular}
\vspace{-2mm}
\caption{Global training hyperparameters of stage 1 pre-training and stage 2 supervised fine-tuning (SFT) for \ourwork.}
\vspace{-2mm}
\label{tab:supp:hyper-global}
\end{table}

\vspace{1mm}
\mypar{Vision Encoder Hyperparameters.} Table~\ref{tab:supp:hyper-vision} presents the specific hyperparameters for each vision encoder integrated into our model. Each encoder is optimized for different input image resolutions and operates with distinct hidden feature dimensions. Following feature extraction, we resize the spatial dimensions of all feature embeddings to $32\times 32$ and concatenate them along the feature channel dimension, producing a unified tensor of shape $32\times 32\times 5120$ tensor. This concatenated representation is then processed by the multimodal MLP projector, which maps the feature channels to match the LLM's hidden dimension of $4096$, ultimately generating $1024$ visual tokens.

\begin{table}[!t]
\centering
  \begin{tabular}{l|c}
    \toprule
    Parameters & Values \\
    \midrule
    CLIP input resolution & $448\times 448$ \\
    CLIP hidden size & 1024 \\
    ConvNeXt input resolution & $1024\times 1024$ \\
    ConvNeXt hidden size & 3072 \\
    EVA-02 input resolution & $1024\times 1024$ \\
    EVA-02 hidden size & 1024 \\
    Grid size & $32\times 32$ \\
    Aspect ratio & square  \\
    Pre-processing & padding \& resizing  \\
    Global hidden size & $5120$ \\
  \bottomrule
\end{tabular}
\vspace{-2mm}
\caption{Hyperparameters for vision encoder designs of \ourwork.}
\vspace{-2mm}
\label{tab:supp:hyper-vision}
\end{table}

\subsection{Training Dataset}

In this section, we detail the datasets utilized in our model training pipeline. For pre-training, we adopt the standard LLaVA-595K~\cite{llava} dataset, following the training protocols established by recent state-of-the-art MLLMs. For supervised fine-tuning, we employ a diverse mixture of datasets from multiple sources.

\vspace{1mm}\mypar{Eagle-1.8M}~\cite{eagle}. Eagle-1.8M represents a comprehensive collection of conversational data aggregated from various specialized datasets, comprising LLaVA-Instruct~\cite{llava} (665K), DocVQA~\cite{mathew2021docvqa} (39K), synDog-EN~\cite{kim2022ocrfree-syndogen} (50K), ChartQA~\cite{masry2022chartqa} (28K), DVQA~\cite{kafle2018dvqa} (25K), AI2D~\cite{kembhavi2016ai2d} (15K), ShareGPT-4V~\cite{chen2023sharegpt4v} (100K), LAION-GPT4v~\cite{laion-gpt4v} (11K), LVIS-Instruct4V~\cite{wang2023lvis-instruct4v} (220K), LRV-Instruct~\cite{liu2023lrv-instruct} (150K), Geo170K~\cite{gao2023geollava} (120K), LLaVAR~\cite{zhang2023llavar} (20K), Visual7W~\cite{zhu2016visual7w} (70K), and Open-Hermes 2.5~\cite{openhermes2-5} (300K). This diverse collection spans a wide spectrum of reasoning scenarios and establishes a robust foundation for our vision-centric reasoning capabilities.

\vspace{1mm}\mypar{VCoT}~\cite{shao2024visualcot}. VCoT comprises a diverse collection of datasets featuring paired bounding box annotations and image-question pairs, including TextVQA\cite{textvqa} (16K), TextCaps~\cite{sidorov2020textcaps} (32K), DocVQA~\cite{mathew2021docvqa} (33K), DUDE~\cite{van2023documentdude} (15K), SROIE~\cite{huang2019sroie} (4K), Birds-200-2011~\cite{wah2011caltechbirds} (10K), Flickr30K~\cite{plummer2015flickr30k} (136K), Visual7W~\cite{zhu2016visual7w} (43K), InfographicsVQA~\cite{mathew2022infographicvqa} (15K), VSR~\cite{liu2023vsr} (3K), GQA~\cite{hudson2019gqa} (88K), and Open images~\cite{openhermes2-5} (43K). 

\vspace{1mm}\mypar{GRIT}~\cite{peng2023kosmos2}. Grounded Image-Text pairs (GRIT) is a large-scale dataset extracted from COYO-700M~\cite{kakaobrain2022coyo-700m} and LAION-2B~\cite{laion-gpt4v}. It is constructed through a pipeline that extracts and links noun phrases and referring expressions in image captions to their corresponding visual regions. Each sample contains an image, caption, extracted noun chunks with corresponding bounding boxes, and two CLIP scores~\cite{radford2021clip} (from ViT-B/32 and ViT-L/14) measuring text-image similarity. Following~\cite{vocot}, we retain 756K samples after filtering out entries with CLIP scores below 0.35.

\vspace{1mm}\mypar{Shikra}~\cite{shikra}. Shikra offers a curated collection of perception-centric datasets specifically designed for object grounding instruction tuning. From its composition, we utilize 326K visual grounding-oriented samples from the RefCOCO-family datasets (RefCOCO, RefCOCO+, RefCOCOg)\cite{refcoco,referitgame}, Visual Genome~\cite{krishna2017visualgenome}, Visual-7w~\cite{zhu2016visual7w}, and Flickr30K~\cite{plummer2015flickr30k}.

\section{Additional Qualitative Visualization}
\label{sec:supp:more_visualization}

\begin{figure*}[!t] 
    \centering
    \captionsetup{type=figure}
    \includegraphics[trim=0 0 0 0, clip=True, width=0.9\linewidth]{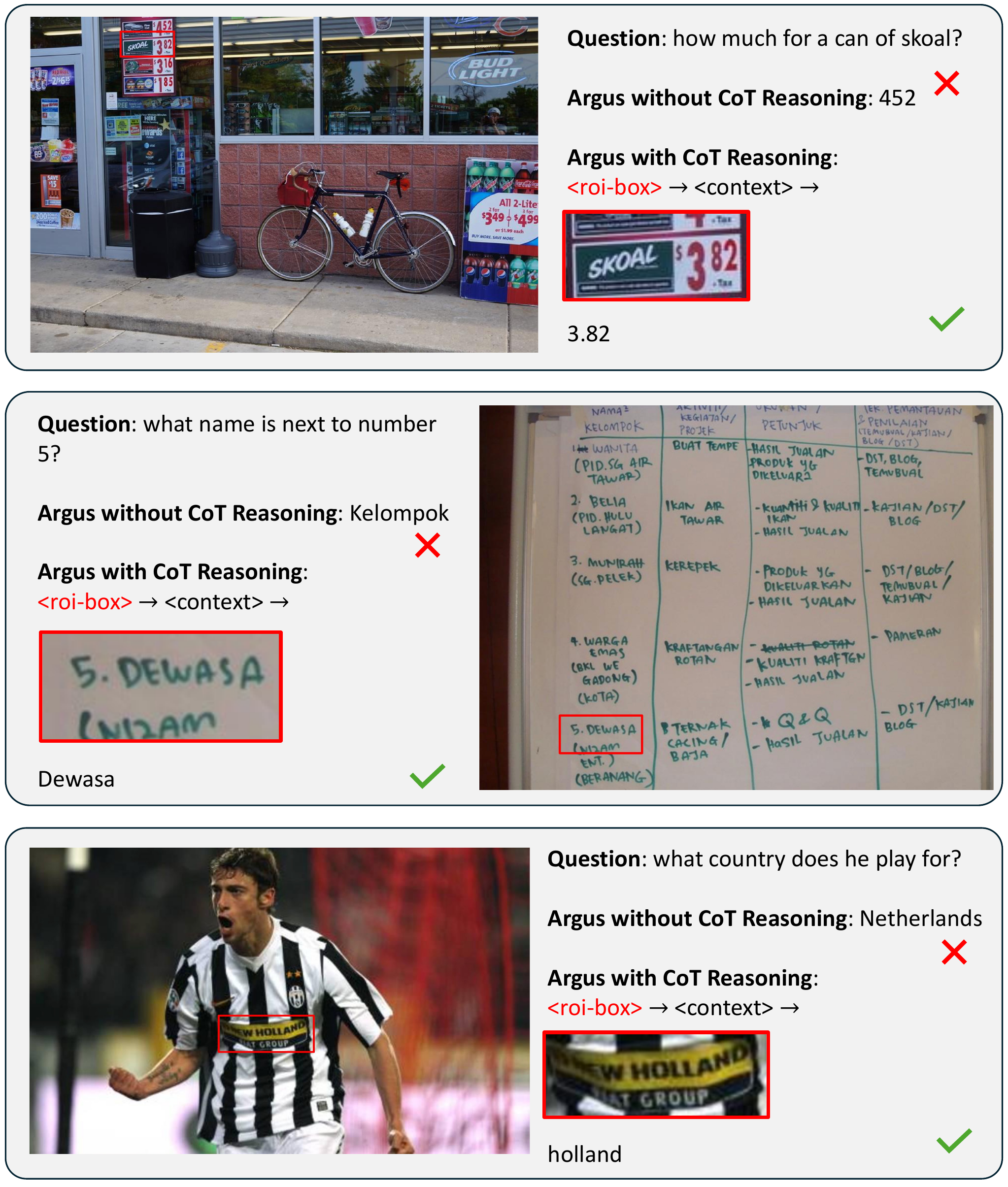}
    \vspace{-2mm}
    \captionof{figure}{\ourwork performance on TextVQA~\cite{textvqa} benchmark, emphasizing on text localization and interpretation in the images.
    } 
    \vspace{30mm}
    \label{fig:supp:more-visualization-01}
\end{figure*}

\begin{figure*}[!t] 
    \centering
    \captionsetup{type=figure}
    \includegraphics[trim=0 0 0 0, clip=True, width=0.9\linewidth]{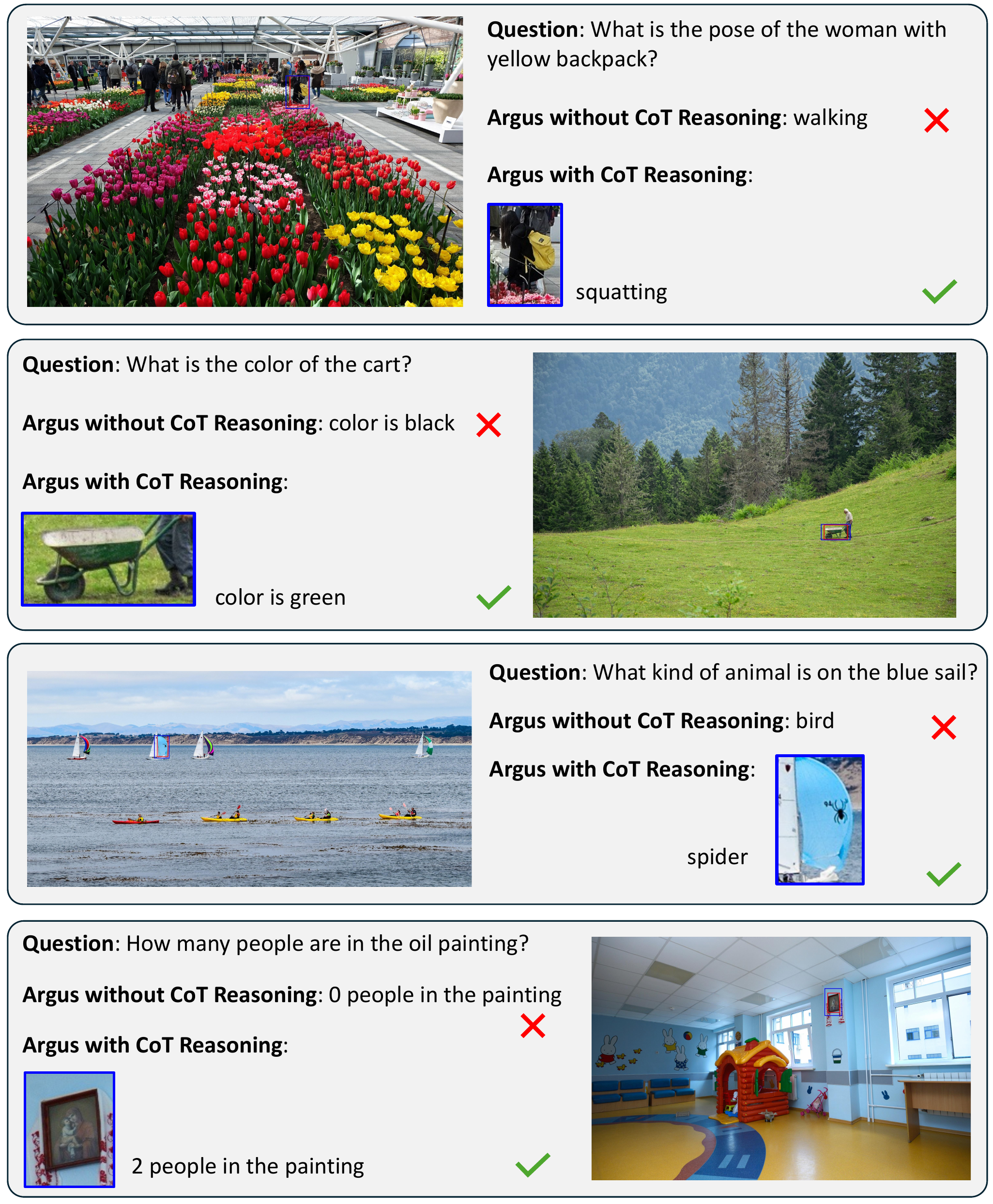}
    \vspace{-1mm}
    \captionof{figure}{\ourwork performance on V-Star~\cite{vstar} benchmark, emphasizing visual perception of objects and regions in complex scenarios. Ground truth bounding boxes are represents in red, and our predicted bounding boxes are in blue.  
    } 
    \vspace{20mm}
    \label{fig:supp:more-visualization-02}
\end{figure*}

We provide additional visualization of the qualitative results on multimodal benchmarks~\cite{masry2022chartqa,vstar} in Figure~\ref{fig:supp:more-visualization-01} and Figure~\ref{fig:supp:more-visualization-02}.

\section{Limitations and Future Work}
\label{sec:supp:limitations}

While we have made substantial progress in exploring the design space of MLLMs for vision-centric reasoning tasks, we acknowledge several limitations in our current approach. This section discusses these limitations and outlines potential directions for future research.

\vspace{1mm}\mypar{Model Capacity.} Our investigation primarily focuses on the design space of visual CoT mechanisms with grounding signals, utilizing the 8-billion parameter Llama3~\cite{dubey2024llama3} model as our LLM decoder backbone. This specific architectural choice may limit the generalizability of our findings. A natural extension of this work would be to evaluate our approach across a spectrum of model scales to validate whether our findings remain consistent in larger architectural configurations.

\vspace{1mm}\mypar{Dataset Complexity.} While \ourwork leverages a diverse combination of multimodal reasoning, visual CoT, and perception/grounding datasets, the current landscape of visual CoT signals remains limited in diversity. Unlike language-based CoT signals, which are abundantly available in internet-scale text corpora and existing language datasets, visual CoT signals are rarely present in large-scale vision-language datasets. Although we have demonstrated significant improvements with the available data, we acknowledge that access to larger-scale, higher-quality visual CoT data would likely yield substantial performance gains and potentially reveal novel emergent capabilities. We believe such data could be derived from existing visual perception datasets or through targeted human annotation efforts, presenting an important avenue for future research.

\vspace{1mm}\mypar{Expanded Vision-centric Tasks Coverage.} While our evaluation of Argus focuses on visual question answering and referring object grounding tasks -- which effectively demonstrate the synergy between perception and reasoning objectives -- we recognize this scope as potentially not comprehensive. In pursuit of developing a truly vision-centric generalist model, a crucial capability would be support for open-world detection tasks. However, given the substantial computational resources and time required for such an investigation, we regard this exploration as future work.

%% file: tables/supp_roi_expansion.tex
\begin{table}[!t]
    \centering
    \fontsize{4pt}{4.8pt}\selectfont
    \setlength\tabcolsep{2pt} % Default value: 6pt
    \renewcommand{\arraystretch}{1.1} % Adjusts the row height
    \scalebox{2.35}{
    \begin{tabular}{r |cccc}
      &
      \multicolumn{2}{c}{Re-encoding} &
      \multicolumn{2}{c}{Re-sampling} \\
      Method &
      \rotatebox{0}{ChartQA} &
      \rotatebox{0}{V-Star}  &
      \rotatebox{0}{ChartQA} &
      \rotatebox{0}{V-Star} \\
      \hline
      0 (\textit{Original}) & \underline{73.1} & 67.3 & \textbf{73.9} & \textbf{67.0}   \\
      + 20\%                & \textbf{73.3} & \underline{67.5} & \underline{73.3} & \underline{66.5}   \\
      + 40\%                & 72.7 & \textbf{68.1} & 73.1 & 64.4   \\
      + 60\%                & 70.1 & 66.5 & 72.5 & 63.3   \\
      + 80\%                & 70.4 & 64.5 & 71.2 & 60.7   \\
   \hline  
   \end{tabular}
}
\vspace{-2mm}
\caption{Impact of RoI context expansion ratios on the performance of re-encoding and re-sampling strategies, evaluated on the ChartQA~\cite{masry2022chartqa} and V-Star~\cite{vstar} benchmarks. Re-encoding demonstrates improved performance with larger context regions, while re-sampling favors the original bounding box size.
}
\vspace{-2mm}
\label{tab:supp:roi-context-expansion}
\end{table}

%% file: tables/supp_LLM_variation.tex
\begin{table*}[!t]
    \centering
    \fontsize{4pt}{4.8pt}\selectfont
    \setlength\tabcolsep{2pt} % Default value: 6pt
    \renewcommand{\arraystretch}{1.1} % Adjusts the row height
    \scalebox{2.21}{
    \begin{tabular}{r |cccccc |ccccc |ccccc}
     \multicolumn{1}{c}{Model} &
     \multicolumn{6}{c}{Vision-Centric Tasks} &
     \multicolumn{5}{c}{Text Understanding} &
     \multicolumn{5}{c}{General Tasks} \\
      
      Method &
      \rotatebox{90}{Avg} &
      \rotatebox{90}{V-Star} &
      \rotatebox{90}{CV-Bench$^\text{2D}$} &
      \rotatebox{90}{CV-Bench$^\text{3D}$} &
      \rotatebox{90}{MMVP} &
      \rotatebox{90}{RealworldQA} &

      \rotatebox{90}{Avg} &
      \rotatebox{90}{ChartQA} &
      \rotatebox{90}{OCRBench} &
      \rotatebox{90}{TextVQA}  &
      \rotatebox{90}{DocVQA}  &
      
      \rotatebox{90}{Avg} &
      \rotatebox{90}{MMMU$^\text{V}$} &
      \rotatebox{90}{MMB} &
      \rotatebox{90}{SEED$^\text{I}$} &
      \rotatebox{90}{GQA}
      \\
      
      \hline
      \rowcolor{gray!20}
      \multicolumn{17}{l}{\textit{Experiments of Argus on other LLMs and Model Sizes}} 
      \\

      Vicuna-7B~\cite{chiang2023vicuna}
      & 57.5 & 64.9 & 61.4 & 59.3 & 41.7 & 60.1 & 65.3 & 70.5 & 52.3 & 67.5 & 71.0 & 61.8 & 38.8 & 69.2 & 75.3 & 63.9
      \\

      Vicuna-13B~\cite{chiang2023vicuna}
      & 60.2 & 66.5 & 64.4 & 62.7 & 43.4 & 64.2 & 70.2 & 74.6 & 56.9 & 74.2 & 75.1 & 63.3 & 39.9 & 72.5 & 75.9 & 65.1
      \\
      
      Llama-8B~\cite{dubey2024llama3}
      & 62.2 & 68.1 & 68.5 & 64.2 & 45.5 & 64.6 & 70.1 & 74.8 & 56.7 & 73.6 & 75.4 & 63.6 & 40.4 & 72.9 & 75.8 & 65.1
      \\
   \hline  
   \end{tabular}
}
\vspace{-3mm}
\caption{{\ourwork supports various choices of LLM backbones. A larger and stronger backbone generally leads to better visual reasoning performance.}}
\vspace{-4mm}
\label{tab:supp:different_llms}
\end{table*}

%% file: tables/supp_multi_roi.tex
\begin{table}[!t]
    \centering
    \fontsize{4pt}{4.8pt}\selectfont
    \setlength\tabcolsep{2pt} % Default value: 6pt
    \renewcommand{\arraystretch}{1.15} % Adjusts the row height
    \scalebox{2.35}{
        \begin{tabular}{l|cc}\hline 
        Model & V-Star & CV-Bench$^{\text{3D}}$ \\\hline
        Argus (single-RoI) & 68.1 & 64.2 \\
        Argus (multi-RoI) & \textbf{78.5} & \textbf{69.6} \\
        \hline
        \end{tabular}
        }
\vspace{-2mm}
\caption{Extension to multiple RoI show great improvement in vision-centric benchmarks~\cite{vstar,cambrian1} where .
}
\vspace{-4mm}
\label{tab:supp:multi_roi}
\end{table}